\documentclass[journal]{IEEEtran}

\usepackage{amsmath}
\usepackage{amsfonts}
\usepackage{amssymb}
\usepackage{amsmath}
\usepackage{color}

\usepackage{graphicx} 

\usepackage{algorithm}
\usepackage{algorithmic}
\usepackage{dcolumn}
\usepackage{hyperref}

\def\R{\mathbb{R}}

\def\X{\mathbf{X}}
\def\x{\mathbf{x}}

\def\d{\mathbf{d}}
\def\y{\mathbf{y}}
\def\w{\mathbf{w}}
\def\r{\mathbf{r}}

\def\e{\mathbf{e}}
\def\p{\mathbf{p}}

\def\s{\mathbf{s}}

\newcommand{\psimplex}{\Delta}

\newcommand{\proba}{\mathbb{P}}
\newcommand{\expectation}{\mathbb{E}}

\newcommand{\Cspace}{\mathcal{C}}

\def\mI{\mathcal{I}}

\newcommand{\rev}[1]{{\color{black} #1}}
\newcommand{\revII}[1]{{\color{black} #1}}
\newcommand{\revIII}[1]{{\color{black} #1}}

\begin{document} 

\title{Greedy methods, randomization approaches and multi-arm bandit algorithms
for efficient sparsity-constrained optimization}
\author{A. Rakotomamonjy~\IEEEmembership{Member,~IEEE}, S. Ko\c{c}o, L. Ralaivola
\IEEEcompsocitemizethanks{
\IEEEcompsocthanksitem AR is with University of Rouen, LITIS Lab. Most of
this work has been carried while he was visiting LIF at Aix-Marseille University. ~\protect Email alain.rakoto@insa-rouen.fr\protect
\IEEEcompsocthanksitem SK is with University of Rouen, LITIS LAB. \protect Email: sokol.koco@gmail.com\protect 
\IEEEcompsocthanksitem LR is with Aix-Marseille University, LIF. \protect Email: liva.ralaivola@lif.univ-mrs.fr\protect
 }
\thanks{This work is supported by Agence Nationale de la Recherche, project
GRETA 12-BS02-004-01.}
\thanks{Manuscript received August2015; revised XXX.}
}

\maketitle
\begin{abstract} 
Several sparsity-constrained algorithms such as 
Orthogonal Matching Pursuit or the Frank-Wolfe algorithm
with sparsity constraints work by iteratively selecting
a novel atom to add to the current non-zero set of variables. This selection step
is usually performed  by computing the gradient
and then by looking for the gradient component
with maximal absolute entry. This step can be computationally
expensive especially for large-scale and high-dimensional
data. In this work, we aim at accelerating these sparsity-constrained
optimization algorithms by exploiting the key observation
that, for these algorithms to work, one only needs the coordinate
of the gradient's top entry. Hence, we introduce
algorithms based on greedy methods and randomization
approaches that aim at cheaply estimating the gradient and
its top entry. Another of our contribution is to cast the
problem of finding the best gradient entry as a best arm
identification in a multi-armed bandit problem. Owing
to this novel insight, we are able to provide a
bandit-based algorithm that directly estimates the top
entry in a very efficient way. Theoretical observations
stating that the resulting inexact Frank-Wolfe or Orthogonal 
Matching Pursuit algorithms act, with high probability, similarly to their exact
versions  are also given.
We have carried out several 
experiments  showing that the greedy deterministic
and the bandit approaches we propose can achieve an
acceleration of an order of magnitude
while being as efficient as the exact gradient
when used in algorithms such as OMP, Frank-Wolfe
or CoSaMP.  
\end{abstract} 

\section{Introduction}

Over the last decade, there has been a large interest in inference
problems featuring data of very high-dimension and a small number of observations. Such problems occur in a wide variety of application domains ranging
from computational biology and text mining to information retrieval and finance.
In order to learn from these datasets, statistical models are frequently designed so as to feature some sparsity properties. Hence, fueled  by
the large interest induced by these application domains, an important amount of research
work in the machine learning, statistics and signal 
processing communities, has been devoted to  sparse learning
and as such, many algorithms have been developed
for yielding models that use only few dimensions of the data.   
To obtain these models, one typically needs to solve a problem
of the form 
\begin{equation}\label{eq:trueprob}
\min_\w   L(\w)   \text{ subject~to }   \|\w\|_0 \leq K 
\end{equation}
where $L(\w)$ is a smooth convex objective function that measures a goodness
of fit of the model,  $\|\w\|_0$ is the $\ell_0$ pseudo-norm that
counts the number of non-zero components of the vector $\w$ and 
$K$ is a parameter that controls the sparsity level. 

One usual approach that has been widely considered is the use of 
a convex and continuous surrogate of the  $\ell_0$ pseudo-norm, namely
an $\ell_1$-norm. The resulting problem is the well-known Lasso
problem \cite{Tibshrani_Lasso_1996} and a large variety of algorithms for its resolution exist,
ranging from homotopy methods \cite{efron_lars} to the Frank-Wolfe (FW) algorithm \cite{jaggi13:_revis_frank_wolfe}. In the same flavour, non-convex continuous penalties  are also common solutions for relaxing  the $\ell_0$ pseudo-norm
\cite{rakotomamonjy2015dc,laporte13:_noncon_regul_featur_selec_rankin}.
Another possible approach is to consider greedy methods that provide
local optimal solution to the problem (\ref{eq:trueprob}). In this last context, a flurry of algorithms have been proposed, the most popular
ones being the \emph{Matching Pursuit} (MP) and the \emph{Orthogonal Matching Pursuit} (OMP) algorithms \cite{mallat_mp,pati93:_orthog_match_pursuit,merhej11:_embed_prior_knowl_within_compr}. 
One common point  of the aforementioned algorithms for solving problem
(\ref{eq:trueprob}) is that they require,
 at each iteration, the computation of the objective function's gradient. For large-scale and high-dimensional setting, computing the gradient at
each iteration may be very time-consuming.

Stochastic gradient descent (SGD) algorithms are now classical methods
for avoiding the computation of the full gradient in large-scale
learning problems \cite{zhang04:_solvin,shalev-shwartz07:_pegas}.
Most of these works have been devoted to smooth composite optimization
although some efforts addressing $\ell_1$-regularized problems exist
\cite{shalev2011stochastic}. Recently, these SGD algorithms have been 
further accelerated through the introduction of variance
reduction methods for  gradient estimation \cite{johnson2013accelerating}.
In the context of problem ~(\ref{eq:trueprob}) where a non-smooth non-convex
constraint appears, very few works have envisaged
the use of stochastic optimization. Nguyen et al. \cite{nguyen14:_linear}
have proposed stochastic versions of a gradient pursuit
algorithm.
Following a similar direction, by exploiting inexact gradient information,
we address the problem of accelerating
sparsity-inducing algorithms such as Matching Pursuit, Orthogonal
Matching Pursuit and the Frank-Wolfe algorithm with an $\ell_1$ ball
constraint.
However, unlike stochastic gradient approaches, the acceleration
we propose leverages on the fact that at each iteration,
these algorithms seek for the gradient's component
that  has the
largest (absolute) value. 

Hence, our main contribution in this paper is to propose  novel
algorithms that allow to efficiently find this top entry of
the gradient. By doing so, our objective is to  design novel
efficient versions of MP, OMP and FW algorithms while keeping intact
all the properties of these algorithms for sparse
approximation. Indeed, this becomes possible
if at each iteration of MP, OMP or FW, our inexact gradient-based
estimation of the component with largest value
is the same as the one obtained with the exact gradient. 

We propose two approaches, the first one based on a greedy deterministic algorithm and the second one based on a randomized method, whose aim is to build an inexact estimation of the gradient so that its top entry in the same as the exact one.
Next, by casting the problem
as a best arm identification multi-armed bandit problem \cite{bubeck2009pure}, we are able to derive an algorithm that directly estimates
the best component of the gradient. Interestingly,
these algorithms are supported by theoretical
evidences that with high probability, they are able
to retrieve the correct component of the gradient. 
As a consequence, we show that MP, OMP and FW algorithms
employing these approaches for spotting 
the correct component of the gradient behave
 as their exact counter part with high probability.

\revIII{This paper is an extended version of the conference paper \cite{rakotomamonjy2015more}. It provides full details on the context of the problem and
proposes a novel key contribution based on multi-arm bandits. In addition, it gives
enlightning insights compared to related works. Extended experimental analysis
also strengthen the results compared to the conference version.
}
The remainder of the paper is organized as follows. Section \ref{sec:sparse}  presents sparsity-constrained algorithms, formalizes our problem
and introduces the key observation on which we build our acceleration strategy.
Section \ref{sec:algo} provides the different algorithms
we propose for efficiently estimating  the extreme gradient component
of interest.
 A discussion with related works is given in Section \ref{sec:discussion}.
Experimental results are depicted in Section \ref{sec:expe} while
 Section \ref{sec:conclusion} concludes the work and presents
different outlooks.

\section{Sparse Learning Algorithm with  Extreme Gradient Component}
\label{sec:sparse}

In this section, we introduce  the sparse learning problem 
we are interested in, as well as some algorithms that are frequently
used for sparse learning. We then point out 
the prominent common trait of those algorithms, namely  
 {\em the extreme gradient component} property,  and discuss how its estimation can be employed for accelerating 
some classes of sparse learning algorithms.

\subsection{Framework}

Consider the problem where we want to estimate a 
relation between a set of $n$ samples gathered in a vector
$\y \in \R^n$ and the matrix $\X \in \R^{n \times d}$. In a sparse
signal approximation problem, $\X$ would be a matrix whose columns are
the elements of a dictionary and $\y$ the target signal,
while in a machine learning problem, the $i$-th row of matrix $\X$,
formalized as $\x_i^\top$, $\x_i \in \R^d$, depicts
the features of the $i$-th example and $y_i$ is the label or target
associated to that example. In the sequel, we denote as $x_{i,j}$ the
entry of $\X$ at the  $i$-th row and $j$-th column.

Our objective is to learn the relation
between $\y$ and $\X$ through a linear model of the data denoted $\X\w$
by looking for the vector $\w$ that solves problem (\ref{eq:trueprob}) 
when the objective function is of the form 
$$L(\w)= \sum_i^n \ell(y_i,g(\w^\top \x_i)),$$ 
where $\ell$ is an individual loss function that measures
the discrepancy between a true value $y$ and its estimation, and
 $g(\cdot)$ is a given differentiable function that
depicts the (potential) non-linear dependence of the loss to $\w$.
Typically,
$\ell$ might be the least-square error function, which leads to
 $L(\w)=\frac{1}{2}\|\y - \X\w\|_2^2$
or the logistic loss and we have $L(\w)=\sum_{i=1}^n \log(1+ \exp\{-y_i \x_i^\top \w\}).$
In the sequel, we present two algorithms that solve problem~\eqref{eq:trueprob} by a greedy method and by a continuous and a convex relaxation
of the $\ell_0$ pseudo-norm, respectively.  

\subsection{Algorithms}

\subsubsection{Gradient Pursuit}

This algorithm is a generalization of the greedy algorithm known
as \emph{Orthogonal Matching Pursuit} (OMP) to  generic loss
functions \cite{blumensath2008gradient_ieee}. It can be directly applied to
our problem given in Equation (\ref{eq:trueprob}). 
Similarly to OMP, the gradient pursuit algorithm is a greedy iterative procedure, which, at each iteration, selects 
the largest absolute  coordinate of the loss gradient. This coordinate is added to the set of already selected
elements and the new iterate is obtained as the solution of the original optimization problem restricted to these selected elements while keeping
the other ones at $0$. The detailed procedure is given
in Algorithm \ref{algo:gp}. 
\begin{algorithm}[t]
\caption{Gradient Pursuit Algorithm \label{algo:gp}}
  \begin{algorithmic}[1]
    \STATE set k = 0, initialize $\w_0 = \mathbf{0}$
\FOR{k=0,1, $\cdots$}
\STATE  $i^\star= \arg \max_i |\nabla L(\w_k)|_i$ \STATE $\Gamma_k= \Gamma_k \cup \{i^\star\}$
\STATE  $\w_{k+1} = \arg \min L(\w)$ over $\Gamma_k$ with $\w_{\Gamma_k^C}= \mathbf{0}$
\ENDFOR  
\end{algorithmic}
\end{algorithm}

While conceptually simple, this algorithm comes with theoretical guarantees
on its ability to recover the exact underlying sparsity pattern of the model, for different types of loss functions \cite{pati93:_orthog_match_pursuit,aravkin2014orthogonal,lozano2011group}. 

Several variations of this algorithm have been proposed ranging from methods exploring new pursuits directions instead of the gradient \cite{blumensath2008gradient_ieee}, to methods making only slight changes to the original one that have strongly impacted the ability of the algorithm to recover signals. 
For instance, the \emph{CoSaMP} \cite{needell09:_cosam} and {GraSP}  \cite{bahmani2013greedy} algorithms select the top $2K$ entries in the absolute gradient,
$K$ being the desired sparsity pattern, optimize over these entries and the already selected $K$ ones, and prune the resulting estimate to be $K$-sparse. In addition to being efficient, these algorithms output  sparse vectors whose distances from the true sparse optimum are bounded.
 
\subsubsection{ Frank-Wolfe algorithm}
\label{sec:fw}

The Frank-Wolfe (FW) algorithm aims at solving the following problem
\begin{equation}
  \label{eq:fw}
  \min_{\w \in C} L(\w) 
\end{equation}
with $\w \in \R^d$, $L$ a convex and differentiable function
with Lipschitz gradient and $C$ a compact subset of $\R^d$.  
The FW algorithm, given in Algorithm \ref{algo:fw}, is a  straightforward procedure that solves problem \ref{eq:fw}
by first iteratively looking for a search direction and then updating the current
iterate. The search direction $\s_k$ is obtained from 
the following convex optimization problem (line~\ref{alg:sk} Alg~\ref{algo:fw})
\begin{equation}
\s_k= \arg \min_{\s \in C} \s^\top \nabla L(\w_k),
\label{eq:s_k}
\end{equation}
which may  be efficiently solved,
depending on the constraint set $C$. For achieving sparsity,
we typically choose $C$ as a $\ell_1$-norm ball,
\emph{e.g.} $
C=\{\w : \|\w\|_1 \leq 1\}$, 
which turns~\eqref{eq:s_k} into 
a linear program.
\begin{algorithm}[t]
\caption{Frank-Wolfe Algorithm \label{algo:fw}}
  \begin{algorithmic}[1]
    \STATE set k = 0, initialize $\w_0=\mathbf{0}$
\FOR{k=0,1, $\cdots$}
\STATE  \label{alg:sk} $\s_k= \arg \min_{\s \in C} \s^\top \nabla L(\w_k)$
\STATE $\d_k= \s_k - \w_k$
\STATE set, linesearch or optimize $\gamma_k\in[0;1]$
\STATE $\w_{k+1}= \w_k + \gamma_k \d_k$
\ENDFOR  
\end{algorithmic}
\end{algorithm}
Despite its simplistic nature, the FW algorithm has been shown to be linearly
convergent \cite{guelat86:_some_wolfes,jaggi13:_revis_frank_wolfe}. 
Interestingly, it can also be shown that convergence is preserved
as long as the following condition holds
\begin{equation}
\s_k^\top \nabla L(\w_k) \leq \min_{\s \in C}  \s^\top\nabla L(\w_k) + \epsilon,
\label{eq:approximate}
\end{equation}
where $\epsilon$ depends on the smoothness of $L$ and the step-size $\gamma_k$.
In general cases where the minimization problem~\eqref{eq:s_k} is expensive to solve,
this condition suggests that an approximate solution $\s_k$ may be sufficient,
provided that the true gradient is available. Similarly, if an
inexact gradient information $\hat \nabla L(\w_k)$ is available, convergence is still
guaranteed under the condition that
$
\s_k^\top \hat \nabla L(\w_k) \leq \min_{\s \in C}  \s^\top\nabla L(\w_k) + \epsilon,
$
and some  conditions relating $\hat \nabla L(\w_k)$ and $\nabla L(\w_k)$.
For instance, if $C$ is a unit-norm ball associated with some norm $\|\cdot \|$
and $\s_k$ a minimizer of $\min_{\s \in C}  \s^\top \hat \nabla L(\w_k)$, i.e.
$\s_k=\arg\min_{\s \in C}  \s^\top \hat \nabla L(\w_k),$ then
in order to ensure convergence, it is sufficient to have 
$\s_k$ so that \cite{jaggi13:_revis_frank_wolfe} 
\begin{equation}
 \label{eq:condition}
\|\hat \nabla L(\w_k) -\nabla L(\w_k)\|_\star \leq \epsilon,
\end{equation}
where $\|\cdot\|_{\star}$ is the conjugate norm associated with $\|\cdot\|$.
\iffalse
\begin{eqnarray}
  \epsilon^\prime  & \geq & 
\|\hat \nabla L(\w_k) -\nabla L(\w_k)\|_\star \\\nonumber
 & =  & \| \nabla L(\w_k) - \hat \nabla L(\w_k)\|_\star \\\nonumber
&\geq & \| \nabla L(\w_k)\|_\star - \|\hat \nabla L(\w_k)\|_\star \\\nonumber
&\geq & \|- \nabla f(\w_k)\|_\star - \|- \hat \nabla f(\w_k)\|_\star \\\nonumber
&= & \max_{\s \in C}  \s^\top(-\nabla L(\w_k)) - \max_{\s \in C}( \s^\top(- \hat \nabla L(\w_k)) \\\nonumber
&\geq & -\min_{\s \in C}  \s^\top(\nabla L(\w_k)) + \min_{\s \in C}( \s^\top(\hat \nabla L(\w_k))
\end{eqnarray}
\fi

 Interestingly, the above equation provides a guarantee on the
convergence on an inexact gradient Frank-Wolfe algorithm, 
but unfortunately, the precision needed on the inexact gradient is given with respect to the exact one.
However, while condition \eqref{eq:condition} is impractical,
it conveys the important idea that inexact gradient computation
may be sufficient for solving sparsity-constrained optimization 
problems.

\subsection{Leveraging from the extreme gradient component estimation}

As stated above, the gradient pursuit algorithm needs  to solve at each iteration the following problem
$
j^\star = \arg \max_j |\nabla L(\w_k)|_j,
$
i.e., the goal is to find at each iteration the gradient's component with the largest absolute value.

In the Frank-Wolfe algorithm, similar situations where finding $\s_k$
corresponds to looking for an extreme component of the (absolute)
gradient, occur. For instance, when the constraint set is  the
$\ell_1$-norm ball $
C_1=\{\w \in \R^d: \|\w\|_1 \leq 1\}$
or the positive simplex constraint
$
C_2=\{\w \in \R^d: \|\w\|_1 = 1, w_j \geq 0\}$
\rev{and we denote 
$$
\s^\star= \arg\min_{\s \in C_1}  \s^\top\nabla L(\w_k)\text{ and } j^\star=\arg\max_j \left|\nabla L(\w_k)|_j\right| ,
 $$
or 
$$
 \s^\star = \arg\min_{\s \in C_2}  \s^\top\nabla L(\w_k) \text{ and } j^\star = \arg\min_j \nabla L(\w_k)|_j ,
 $$
then $\s^\star= \e_{j^\star}$, with $\e_{j^\star}$ 
the $j^\star$-th canonical vector of $\R^d$.}
Hence, as in Matching Pursuit, OMP or the gradient pursuit
algorithm, for specific choices of $C$, we are not interested
in the full gradient, nor in its extreme values, but only on the coordinate of the gradient component with the smallest, the largest
 (absolute) value.

Our objective in this paper is to leverage on these observations for
accelerating sparsity-inducing algorithms which look for one extreme
component of the gradient. In particular, we are interested in
situations where the gradient is expensive to compute and we aim at
providing computation-efficient algorithms that produce either
approximated gradients whose extreme component of interest is the same
as the exact gradient's one, or identify the extreme component without
computing the whole (exact) gradient.

\section{Looking for the extreme gradient component}
\label{sec:algo}

This section formalizes the problem of identifying the
extreme gradient component and provides different
algorithms for its resolution. We first introduce 
a greedy approach, then a randomized one and finally exhibit
how this problem can be cast as best arm identification
in a multi-armed bandit framework.

\subsection{The problem}

As  mentioned in Section \ref{sec:sparse}, we are interested in learning problems
where the objective function is of the form 
$$
\sum_i \ell( y_i, g(\w^\top \x_i)).
$$
The gradient of our objective function is thus given by
\begin{align}
\nabla L(\w)& = \sum_i \ell^\prime(y_i, g(\w^\top\x_i))) g^\prime(\w^\top\x_i) \x_i = \X^\top\r  
\end{align}
where $\r \in \R^n$ is the vector whose $i$-th component is
$\ell^\prime(y_i, g(\w^\top\x_i))) g^\prime(\w^\top\x_i)$.
\rev{This particular form  entails that the gradient may be computed iteratively. Indeed, 
the sum $\nabla L(\w)= \sum_{i=1}^n \x_i r_i$ is invariant to 
the order according to which index $i$ decribes the set $[1,...,n]$.
This makes possible the computation, at each iteration $t$, of an approximate gradient $\hat \nabla L_t$.
Denote  $\mI_t$  the first
$t$ indices used for computing the sum and $i_{t+1}$ the 
$(t+1)$-th index, then, at iteration $t$,  $\hat \nabla L_{t}= \sum_{i \in \mI_{t}} \x_i r_i $
and
\begin{equation}\label{eq:accumulation} 
\hat \nabla L_{t+1} =  \hat \nabla L_{t} +  \x_{i_{t+1}} r_{i_{t+1}}. \end{equation}
}
According to this framework, our objective is then to find an efficient way
for computing an approximate gradient $\hat \nabla L_t$ for which the desired
extreme entry is equal to the one of the exact gradient. \rev{Note that 
the computational cost of the exact gradient  $\X^\top\r$ is 
about $O(nd)$ and a naive computation of the residual $\r$ has
 the same cost, which leads to a global complexity of 
$O(2nd)$. However, because $\w$ is typically a sparse vector in Gradient
Pursuit or in the Frank-Wolfe algorithm, we compute
the residual vector $\r$ as  $\r=\y - \X_\Omega \w_\Omega$
where $\Omega$ are the indices of the non-zero elements of the $k$-sparse vector $\w$. 
Hence, computing $\r$ has only a cost of $O(nk)$ 
and it does not require a full pass on all the elements of 
$\X$. The main objective of our contributions is to estimate
the extreme entry of $\X^\top \r$ with algorithms having a complexity
lower than $O(nd)$.
} 
We propose
and discuss, in the sequel, three different approaches. 

\subsection{Greedy deterministic approach}

\begin{algorithm}[t]
    \renewcommand{\algorithmicrequire}{\textbf{Input:}}
    \renewcommand{\algorithmicensure}{\textbf{Output:}}
\caption{Greedy deterministic algorithm to compute
$\hat \nabla L$ }
  \begin{algorithmic}[1]
\label{algo:greedy}
    \REQUIRE: $\r$, $\{\|\x_i\|\}$, $\X$
\STATE [values, indices]=sort $\{|r_i| \|\x_i\| \}_i$ in decreasing order
\STATE $\hat \nabla L_t = \mathbf{0}$, $t= 0$
\REPEAT
\STATE $i=$indices$(t+1)$ 
\STATE $\hat \nabla L_{t+1}= \hat \nabla L_{t} + \x_i r_i $
\STATE $t = t+1$
\UNTIL{stopping criterion over $\hat \nabla L_{t+1}$ is met }
\ENSURE $\hat \nabla L_{t+1}$
\end{algorithmic}
\end{algorithm}

The first approach we propose is a greedy approach which, at iteration
$t$, looks for the best index $i$ so that $r_i \x_i$  optimizes  
a criterion depending on $\hat \nabla L_t$ and $\nabla L$.

Let $\mI_t$ be the set of  indices of the examples chosen in the first $t$ iterations for computing $\hat \nabla L_t$.
At iteration $t+1$, our goal is to find the example $i^\star$ that is the solution of the following problem:
\begin{align}
i^\star &=  \operatorname{argmin}\limits_{i\in \{1,\ldots,n\} \backslash \mI_t}
\|\nabla L -\hat \nabla L_{t} - \x_i r_i\|,
\end{align}
with $\hat \nabla L_{t+1}= \hat \nabla L_{t}+ \x_i r_i$.
The solution of this problem is thus the vector product
$r_i\x_i$ that has minimal norm difference compared to the current gradient estimation
residual $\nabla L - \hat \nabla L_{t}$. While simple, this solution
cannot be computed because $\nabla L$ is not accessible. 
Hence, we have to resort to an approximation based on
the following equivalent problem: 
\begin{align}
\label{eq:approxgraddiff}
i^\star &= \operatorname{argmax}\limits_{i\in \{1,\ldots,n\} \backslash \mI_t}  - \|\nabla L - \hat \nabla L_{t+1}\| \\
 &= \operatorname{argmax}\limits_{i\in \{1,\ldots,n\} \backslash \mI_t} \|\nabla L - \hat \nabla L_t\| - \|\nabla L - \hat \nabla L_{t+1}\| \nonumber
\end{align}
where we use the fact that $\|\nabla L - \hat \nabla L_t\|$ does
not depend on $i$.
By upper bounding the above  objective value, one can derive the best
choice of $r_i \x_i$ that achieves  the largest variation of the
gradient estimation norm residual. Indeed, we have
\begin{equation}
\|\nabla L - \hat \nabla L_t\| - \|\nabla L - \hat \nabla L_{t+1}\| \leq \| - \hat \nabla L_t  + \hat \nabla L_{t+1}\| \notag
\end{equation}
The right-hand side of this equation can be further simplified
\begin{align}\label{eq:approxgraddiff2}
 \| \hat \nabla L_{t+1} - \hat \nabla L_t\| 
& =\| \hat \nabla L_t - \x_{i} r_{i}  - \hat \nabla L_t\| \notag \\
& =\| \x_{i} r_{i} \| = \| \x_i\| | r_{i}|. 
\end{align}

Equation \eqref{eq:approxgraddiff2} suggests that the index $i$ that should be chosen at iteration $t+1$ is the one with the largest absolute residual weighted by its example norm.
Simply put, in the first iteration, the algorithm chooses the index
that leads to the largest value $\| \x_i\| | r_{i}|$  , in the second one, it selects the second best, and so forth.
That is, the examples are considered in a decreasing order with respect  to the weighted value of their residuals. Pseudo-code of the approach 
is given in Algorithm \ref{algo:greedy}.

Note that for this method, at iteration $t=n$ we recover the exact gradient,
$$
\hat \nabla L_n =  \nabla L.
$$
Hence, when $t=n$, we are assured to retrieve
the correct extreme entry of the gradient at the expense of extra
computations needed for running this greedy approach compared to
a plain computation of the full gradient.  On the other hand,
if we stop this gradient estimation procedure before
$t=n$, we save computational efforts at the risk of missing the correct extreme
entry.

\subsection{Matrix-Vector Product as Expectations}
\label{sec:mvprod}

\begin{algorithm}[t]
    \renewcommand{\algorithmicrequire}{\textbf{Input:}}
    \renewcommand{\algorithmicensure}{\textbf{Output:}}
\caption{Computing $\hat \nabla L$ as an empirical average }
  \begin{algorithmic}[1]\label{algo:empirical}
    \REQUIRE: $\r$, $\|\x_i\|$,$\X$
\STATE build $\p\in\psimplex_n^*$ \emph{e.g} $p_i \propto \frac{1}{n}$ 
or $ p_i\propto |r_i|\|\x_{ i}\|$
\REPEAT
\STATE random draw $i \in 1,\cdots,n$ according to $\p$ 
\STATE $\hat \nabla L_{t+1}= \hat \nabla L_{t} + \frac{1}{p_i}\x_i r_i $
\STATE $t = t+1$
\UNTIL{stopping criterion over $\hat \nabla L_{t+1}$ is met }
\ENSURE $\hat \nabla L_{t+1}$
\end{algorithmic}
\end{algorithm}

The problem of finding the extreme component of the gradient can also be addressed from the point of view of randomization, as described
in Algorithm \ref{algo:empirical}.

The approach consists in considering the computation
of $\X^\top \r$  as an the expectation of a given random variable. 
Recall that $\X$ is composed of the vectors $\{\x_i^\top\}_{i=1}^n$
and $\x_i \in \R^d$.
Hence, the matrix-vector product $\X^\top\r$ can be rewritten:
\begin{equation}
\label{eq:decomposition}
\X^\top\r=\sum_{i=1}^{n}r_i\x_{i}.
\end{equation}

From now on, given some integer $n$, $\psimplex_n^*$ denotes the interior of the probabilistic  simplex of size $n$:
 \begin{equation}
 \label{eq:psimplex}
 \psimplex_n^*\doteq\left\{\p=[p_1\cdots p_n]:\sum_{i=1}^np_i=1,\;p_i> 0,i=1,\ldots,n\right\}.
 \end{equation}

 For any element $\p=(p_i)_{i\in[n]}\in\psimplex_n^*$ we introduce a random vector
 $C$ that takes value in the  set
 $$\Cspace\doteq\left\{\mathbf{c}_i\doteq r_i\x_{i}/p_i:i=1,\ldots,n\right\}$$
 so that $ \proba(C=\mathbf{c}_i)=p_i$.
 This way,
 \begin{equation}\label{eq:expectationC}
 \mathbb{E} C = \sum_{i=1}^np_i\mathbf{c}_i=\sum_{i=1}^np_ir_i\x_{i}/p_i=\sum_{i=1}^n r_i\x_{i}=\X^\top\r
 \end{equation}
Hence, if $C_1,\cdots,C_s$ are independent copies of $C$ and $\hat{C}^s$ is
defined as:
$
\hat{C}^s\doteq \frac{1}{s}\sum_{i=1}^s C_i
$
then
$
\expectation\hat{C}^s=\X^\top\r.
$
$\hat{C}^s$ is thus an estimator of the matrix-vector product that we are interested in \emph{i.e} the gradient of our objective function.

According to the above, a relevant approach for estimating  the extreme component of the gradient is to randomly sample  $s$ copies of $C$, to average them
and then to look for the extreme component of this estimated gradient.

Interestingly, this approach based on randomized matrix multiplication
can be related to our deterministic approach. Indeed, a result given by
\cite{drineas2006fast} (Lemma 4) says that the element $\p\in\psimplex_n^*$ that minimizes  $ \expectation[\|\X^\top\r - \hat{C}^s\|^2_F]$ is such that
\begin{equation}\label{eq:best}
p_i\propto |r_i|\|\x_{i}\|.
\end{equation}
It thus suggests that vectors $c_i$ of large values of $|r_i|\|\x_{i}\|$ have higher probability to be sampled. This resonates with the greedy
deterministic approach in which vectors $\x_ir_i$ are accumulated
in the order of the decreasing values of $|r_i|\|\x_{i}\|$.

\subsection{Best arm identification and multi-armed bandits}

The two preceding approaches seek at approximating the
gradient, at a given level of accuracy that has yet to be defined,
and then at evaluating the coordinate of its extreme component.

Yet, the problem of finding the extreme component coordinate of a gradient vector
obtained from matrix-vector multiplication can also be posed
as the problem of finding the best arm in a multi-armed bandit
problem. In a nutshell, given a slot machine with multiple arms, the goal in the bandit problem is to find the arm that maximizes the expected reward or minimizes the loss.
For this, an iterative procedure is used, where at each step, a forecaster selects an arm, based on his previous actions, and receives a reward or observes a loss.
Depending on how the reward is obtained, the problem can be stochastic
(the reward/loss is drawn from a probability distribution) or non-stochastic.
Bubeck et al. \cite{bubeck2009pure} propose an extensive review of these methods for various settings.

We cast our problem of finding the extreme gradient component as a
best arm identification problem as follows. In the remainder of this section, we suppose that we look for a minimum gradient component and thus we look for the arm with minimal loss instead of maximal reward.
We consider that the arms are the components of the gradient 
(we  have $d$ arms) and at each pull of a given arm, we observe
a loss that is built from  a term of the gradient matrix-vector
multiplication $\X^\top r$, as made clear in the sequel.

In a stochastic setting, we consider a similar framework
as the one we described  in Section \ref{sec:mvprod}.
We model the loss obtained from the $k$-th pull of 
 arm $j \in [1,\cdots,d]$ as a random variable $V$, independent of $k$, that takes value in the set
$$\left\{v_{j,i} \doteq r_i x_{i,j}/p_i:i=1,\ldots,n\right\}$$
so that $ \proba(V=v_{j,i})=p_i$. From this definition, the expected
loss of arm $i$ is 
$$ \mathbb{E} V = \sum_{i=1}^np_i v_{j,i} = \sum_{i=1}^n r_i x_{i,j} = 
(\X^\top\r)_j,
 $$
which is the $j$-th component of our gradient vector. 
 In this setting,
given a certain number of pulls, one pull providing a
realization of the random variable $V$ of the chosen arm,
the objective of a best arm identification algorithm is to provide
recommendation about the arm with minimal
expected loss, which in our case is the 
coordinate of smallest value in 
the $d$-dimensional vector $\X^\top\r$.

Several algorithms for identifying this best
arm have been provided in the literature. Most of them
are built around an empirical average loss statistic.
This latter can be computed, after $s$ pulls of 
an arm $j$, as
$
\hat V_{j,s} = \frac{1}{s} \sum_{t=1}^s v_{j,i(k,j)}
$,
where $i(k,j)$ is the index $i$ drawn at
the $k$-th pull for arm $j$.
Some of the most
interesting algorithms are the \emph{successive reject} \cite{audibert2010best}
and \emph{successive halving} \cite{karnin2013almost} algorithms which, given a fixed budget
of pulls, iteratively discard after some predefined
number of pulls (say $s$) the worst arm or the worst-half arms, respectively,
according to the values $\{\hat V_{j,s}\}_{j=1}^d$.  
These approaches are relatively simple and the \emph{successive halving}  approach is depicted in detail in Algorithm \ref{algo:sh}. 
They directly provide some guarantees on the
probability to correctly estimate the extreme component of the gradient.
For instance, for a fixed budget of pulls $T$, under some minor
and easily satisfied conditions on $\{v_{j,i}\}$, the \textit{successive
halving} algorithm correctly identifies the minimum gradient component with
probability at least $1-3 \log_2 d \cdot \exp(-\frac{T}{8 H_2 \log_2 d})$
where $H_2$ is a problem-dependent constant (the larger
$H_2$ is, the harder the problem is).

However, these two algorithms have the drawbacks to work on individual
entries $x_{i,j}$ of the matrix $\X$. Hence, overload due to single memory accesses
compared to those needed for accessing chunks of memory may hinder
the computational gain obtained by identifying the component
with minimal gradient value without computing the full gradient.
 
For the \textit{successive halving} algorithm, one way of overcoming this issue is to consider \emph{non i.i.d} sampling
of the arm's loss. As such, we consider that at each iteration, the losses
generated by pulling the remaining arms come from the same component $j$
of the residual. \rev{This approach has 
the advantage of working on the full vector $\x_i r_i$, 
allowing thus  efficient memory caching, instead of
 on individual elements of $\X$.

Let $\mathcal{A}_{s^\prime}$ and $\mathcal{A}$ denote the sets
of components $i$ that have been drawn after 
$s^\prime$ and the subsequent $s^\dagger$ pulls such
that $s = s^\prime + s^\dagger$ respectively, then
the loss for arm $j$ after $s$ pulls can be defined as
\begin{equation}\label{eq:noniid}
\hat V_{j,s} 
= \hat V_{j,s^\prime} + \sum_{i \in \mathcal{A}} \frac{r_i x_{i,j}}{p_i},
\end{equation}
where $\hat V_{j,0}=0$ and the set
$\mathcal{A}$ is the same for all arms.
 In practice, as implemented in the numerical simulations we provide, each component $i$
of $\mathcal{A}$ is randomly chosen according to a uniform distribution over $1,\cdots,n$.
The \emph{non-iidness} of the stochastic process
comes from that the arm losses
are dependent through the set $\mathcal{A}$. However,
this does not hinder the fact that empirical loss $\hat V_{j,s}$ is still 
a relevant estimation of the expected loss of arm $j$.}

\begin{algorithm}[t]
\caption{Successive Halving to find the minimum gradient component\label{algo:sh}}
  \begin{algorithmic}[1]
    \STATE \textbf{inputs:} $\X$, $\r$, set $T$ the budget of pulls
\STATE Initialize $S_0= [d\,]$
\STATE $\hat V_{j,0} = 0, \forall j \in S_0$
\FOR{$\ell$= 0,1, $\cdots, \lceil log_2(d) \rceil$ -1}
\STATE Pull each arm in $S_\ell$ for $r_\ell = 
\lfloor \frac{T}{|S_\ell| \lceil \log_2(n) \rceil}\rfloor
$ additional times and
compute resulting loss (\textit{non-iid}) $\hat V_{\cdot,R_\ell}$ in Equation (\ref{eq:noniid}) or (\textit{non-stochastic}) $v_{\cdot,R_\ell}$ in Equation (\ref{eq:banditnonstoch}) with $R_\ell = \sum_{u=0}^\ell r_u$
\STATE Sort arms in $S_\ell$ by increasing value of losses
\STATE Define $S_{\ell+1}$ as the $\lfloor|S_\ell|/2 \rfloor$ indices of arms with smallest values
\ENDFOR  
\STATE \textbf{output:} index of the best arm
\end{algorithmic}
\end{algorithm}

Non-stochastic  best arm identification has been barely studied and
only a very recent work has addressed this problem \cite{jamieson2015non}. In this latter
work, the main hypothesis about the non-stochasticity of the loss
is that they are assumed to be generated before the game starts. This is
exactly our situation since \rev{before each estimation
of the minimum gradient component, all
losses are given by $x_{i,j} r_i$ and thus 
can be computed beforehand}. In this
non-stochastic setting \cite{jamieson2015non}, the framework is that the 
$k$-th pull of an arm $j$ provides a loss $v_{j,k}$ and
the objective of the bandit algorithm is to identify
$
\arg \min_{j} \lim_{k \rightarrow \infty} v_{j,k},
$
assuming that such limits exist for all $j$. 
Again we can fit our problem of finding the extreme gradient
component (here the minimum) into this framework by 
defining the loss for a given arm at pull $k$ as 
\begin{equation}\label{eq:banditnonstoch}
 v_{j,k} = \left \{
\begin{array}{ll} 
0 & \text{ if } k=0 \\
v_{j,k-1} + r_{\tau_k} x_{\tau_k,j}  & \text{ if } 1 \leq k \leq n \\
v_{j,k-1}   & \text{ if }  k > n \\
\end{array}
\right.
\end{equation}
where $\tau$ is a predefined or random permutation of 
the rows of the vector $\r$ and the matrix $\X$, and \rev{$\tau_k$
its $k$-th entry.} 
In practice, we choose  $\tau$ to be the same for all the arms
for computational reasons as explained above, but in theory this 
is not necessary \cite{jamieson2015non}.
According to this loss definition, we have 
 $$
 \lim_{k \rightarrow \infty} v_{j,k} = \sum_{k=1}^n  r_{\tau_k} x_{\tau_k,j}  
=  \sum_{k=1}^n  r_{k} x_{k,j} = (\X^\top\r)_j.
 $$
Hence, an algorithm that recommends the best arm after a given number
of pulls, will return the index of the minimum component in our
gradient. Interestingly, the algorithm proposed by 
Jamieson et al. \cite{jamieson2015non} for solving the non-stochastic
best arm identification problem is also the one used in
the stochastic setting namely the \emph{successive halving}
algorithm (Alg. \ref{algo:sh}). This algorithm can be
shown to work as is despite the dependence between  arm losses. 
Indeed, each round-robin pull of the surviving arms can have dependent values,
 and as long as the algorithm does not adapt to the observed losses 
during the middle of a round-robin pull  \cite{jamieson2015non}.

As already mentioned, for a fixed budget $T$ of pulls, this
\emph{successive halving}
 bandit algorithm  comes with theoretical guarantee in its ability to identify the best arm. 
In the stochastic case, the probability of success depends on the number of
arms, the number of pulls $T$ and on a parameter denoting the hardness
of the problem (see Theorem 4.1 in \cite{karnin2013almost}).
In the non-stochastic case, the budget of pulls needed for
guaranteeing the correct recovery of the best arm essentially
depends on a function $\gamma_j(k)$ such that
$
|v_{j,k} - \lim_{k \rightarrow \infty} v_{j,k}| \leq \gamma_j(k)
$
and on a parameter denoting  the gap between  
any of $(\X^\top \r)_j$ and the smallest component of  
$\X^\top \r$ which is not accessible unless we compute the
exact gradient.

One may note the strong resemblance between the matrix-vector
product approximation  as given in 
\eqref{eq:expectationC} and the \emph{non-iid} bandit strategy,
 \rev{as in this latter setting, 
we consider the full vector $\x_i r_i$ to
compute $\hat V_{\cdot,s}$ for all remaining arms. 
This \emph{non-iid} strategy can also be related
to the non-stochastic setting if we choose
$\tau$ as a random permutation of $1,\ldots,n$.
} 
In addition, in the non-stochastic
bandit setting, we can recover the greedy deterministic approach
if we assume that the permutation $\tau$ defines a re-ordering
of $\|\x_i\||r_i|$ in decreasing order, then the accumulation
given in Equation \eqref{eq:banditnonstoch} is exactly the one given in the greedy deterministic
approach. \rev{This is the choice of $\tau$  we have considered in our experiments.} Multi-armed bandit  framework
and the gradient approximation approaches use thus
similar ways for computing the criteria used for
estimating the best arm. The main difference resides
in the fact that with multi-armed bandit,
one is directly provided with the 
estimation of the best arm.

\subsection{Stopping criteria}

In the greedy deterministic and randomized methods introduced in this
section, we have no clues on how many elements $ r_i \x_i$ have to be
accumulated in order to achieve a sufficient approximation of the
gradient or in the multi-armed bandit approach how many pulls we need
to draw. Here, we discuss two possible stopping criteria for the
non-bandit algorithms: one that holds for any approach and a second
one that holds only for the Frank-Wolfe algorithm in the deterministic
sampling case. Discussion on the budgets that needs to be allocated to
the bandit problem is also provided.

\subsubsection{Stability condition}
For the sake of simplicity, we limit the exposition to the search
of the smallest component of the gradient, although
the approach can be generalized to other cases.

Denote by $j^\star$ the coordinate such that $j^\star=\arg\min_j \nabla L(\w_k)|_j$
and let $T_s$ be the maximal number of iterations or samplings allowed
for computing the inexact gradient (for instance, in the greedy deterministic approach, $T_s=n$). Our objective
is to estimate $j^\star$ with the fewest number $t$ of iterations.
For this to be possible, we make the hypothesis that there exists
an iteration $t_1$, $t_1 \leq T_s$, and
$$
j^\star=\arg\min_j \hat \nabla L_{t}(\w_k)|_j  \quad \forall t :   t_1 \leq t \leq T_s
$$ 
in other words, we suppose that starting from a given number of iterations
$t_1$, the gradient approximation is sufficiently accurate so that the updates of the gradient will leave 
the minimum coordinate unchanged. Formally, this condition means
that $\forall t \in [t_1,\cdots,n]$, we have
$$
\Big[\hat \nabla L_t + \sum_{u=0}^{T_s-t} U_{t+u} \Big]_{j^\star} 
\leq \Big[\hat \nabla L_t + \sum_{u=0}^{T_s-t} U_{t+u} \Big]_{j} 
\quad \forall j \in [1,\cdots,d] .
$$
where each $U_{t+i}= r_{i(t+u)} \x_{i(t+u)}$, $i(t+u)$ being
an index of samples that depends how the greedy or randomized 
strategy considered.
However, checking the above condition is as expensive as 
computing the full gradient, thus we propose an
estimation of $j^\star$ based on an approximation of this inequality,
by truncating the sum
to few iterations on each side. Basically, this consists in
evaluating $j^\star$ at each iteration and checking
whether this index has changed over the last $N_s$ iterations.
 We refer to this criterion 
as the \textit{stability criterion}, parametrized
by $N_s$.

\subsubsection{Error bound criterion}
In Section \ref{sec:fw}, we discussed that the convergence of the Frank-Wolfe inexact algorithm can be guaranteed as long as the norm difference between the approximate gradient and the exact one could be upper-bounded by some quantity $\epsilon$. Formally, this means that the iterations over gradient approximation can be stopped as soon as
$\|\hat \nabla L_t - \nabla L \|_\infty \leq \epsilon$,
where $\epsilon$ depends on the curvature of the function $L(\w)$ \cite{jaggi13:_revis_frank_wolfe}. In practice, the criterion 
$\|\hat \nabla L_t - \nabla L \|_\infty$ cannot be computed as it depends on the exact gradient but it can  be upper-bounded by a term that is accessible. For the greedy deterministic approach, by norm equivalence, we have 
\revII{
\begin{equation}\label{eq:errorcrit}
   \|\hat \nabla L_t - \nabla L \|_\infty =
\Big\|\sum_{i \not \in \mI_t} \x_i r_i \Big \|_\infty \leq 
\sum_{i \not \in \mI_t} \|\x_i\|_\infty |r_i| \leq \epsilon
 \end{equation}
 Hence if the norms $\{\| \x_i\|_\infty\}$ have been precomputed beforehand, this criterion can be easily evaluated at each gradient update iteration.  
}

\subsubsection{Pull budget for the bandit}

In multi-armed  bandit algorithms, one typically specifies the number of pulls $T$ available for estimating the best arm. As such, $T$
can be considered as  hyperparameter of the algorithm. 
A possible strategy for removing the dependency of the
bandit algorithm to this pull budget, is to use the
\textit{doubling trick} \cite{jamieson2015non}, which consists in running the algorithm with a small value of $T$ and then repeatedly doubling it until some stopping criterion is met. The algorithm can then
be adapted so as to exploit the loss computations that have already been
carried from iteration to iteration.
However, this strategy needs a stopping criterion for the \emph{doubling
trick}. According to Theorem $1$ in \cite{jamieson2015non}, there
exists a lower bound of pulls for which the algorithm is guaranteed
to return the best arm. Hence, the following heuristic can
be considered: if  $T^\prime$ and $2T^\prime$ number of pulls return the same best arm, then we conjecture that the proposed arm is indeed the best one.
One can note that this idea is similar to the above-described 
stability criterion.
While this strategy is appealing, in the experiment, we have just fixed the
budget of pulls $T$ to a fixed predefined value.

\section{Discussion}
\label{sec:discussion}

This section provides comments and discussions of the
approaches we proposed compared to existing works.

\subsection{Relation and gains compared to OMP and variants}

Several recent works on sparse approximation have 
proposed theoretically founded algorithms. These works
include OMP \cite{pati93:_orthog_match_pursuit,tropp07:signalrecov},
greedy pursuit \cite{blumensath2008gradient_ieee,tropp06:SSA}, CoSaMP \cite{needell09:_cosam} and several others like \cite{shalev-shwartz10:_tradin_optim}. 
Most of these algorithms make use of the 
top absolute entry of the gradient vector at each iteration. The work presented
in this paper is  strongly related to these ones as we share the same
quest for the top entry.
Indeed, the proposed methodology provides tools that can be applied to
many sparse approximation algorithms including the aforementioned ones.
What makes our work novel and compelling is that at each iteration, the
gradient is computed with as few information as possible. 
If the stopping criterion for estimating this gradient 
is based on a maximal number of samples --- \textit{e.g}. we are interested in constructing the best approximation of the gradient from only 20\% of the samples---, our approach can be interpreted as a method for computing the gradient on a limited budget.
Hence, the proposed method allows  to obtain a gain in the computational time needed for the estimation of the gradient.
On the downside, if other stopping criteria are used (alone or jointly with the budget criterion), this gain may be partly impaired by further computations needed for their estimation.
As an example, the \emph{stability} criterion induces a $O(d)$ overhead at each iteration due to the max computation.

\subsection{Relation with other stochastic MP/FW approaches}

\rev{
Some prior works from the literature are related to the approaches we
have proposed in the present paper. 
Chen et al. \cite{chen2011stochastic} have recently introduced
a stochastic version of a Matching Pursuit algorithm in a 
Bayesian context. Their principal contribution was to define
some prior distribution over each component of the vector
$\w$ and then to sample over this distribution so as to estimate
$\w$. In their approach, the sparsity pattern related to Matching Pursuit
is controlled by the prior distribution which is assumed
to be a mixture of two distributions one of which induces sparsity.
While this approach is indeed stochastic, it strongly differs
from ours in the way stochasticity is in play. As we will
discuss in the next subsection, our framework
is more related to stochastic gradient than to the
stochastic sampling of Chen et al. 

Stronger similarities with our work 
appear in the work of Peel et al. \cite{peel2012matching}.
Indeed, they propose to accelerate the atom selection
procedure by randomly selecting a subset of atoms
as well as a subset of example for computing $\X^\top\r$.
This idea is also the base of our work. However,
an essential difference appears as we do
not select a subset of atoms. By doing so, we are
ensured not to discard the top entry of $\X^\top\r$
and thus we can guarantee for instance that  
our bandit approaches are able to retrieve this top
entry with high probability given enough budget
of pulls.

Stochastic variants of the Frank-Wolfe algorithm
have been recently proposed by Lacoste-julien
et al.  \cite{lacoste-julien13:_block_coord_frank_wolfe_struc_svms}
and Ouyang et al. \cite{ouyang2010fast}. These works
are mostly tailored for solving large-scale SVM optimization problem
and do not focus on sparsity.
}

\subsection{About stochasticity}

The randomization approach for approximating the gradient, introduced in Section \ref{sec:mvprod},
involves random sampling of the columns. In the extreme situation
where only a single column  $i$ is sampled, we thus have~
$\hat \nabla L = \x_ir_i,
$
and the method we propose boils down to a stochastic gradient
method. In the context of sparse
greedy approximation, the first work devoted to stochastic gradient approximation has been recently released \cite{nguyen14:_linear}. Nguyen et al. \cite{nguyen14:_linear} show that
their stochastic version of the iterative hard thresholding algorithm,
or the gradient matching pursuit algorithm which aim at greedily solving
a sparse approximation problem with arbitrary loss functions, behave properly in expectation.

The randomized approach we propose in this work goes beyond the
stochastic gradient method for
greedy approximation, since it also provides a novel approach for
computing stochastic gradient. Indeed, we differ from
their setting in several important aspects:
\begin{itemize}
\item First, in our stochastic gradient approximation,
we always consider a  number of samples larger than $1$. As such,
we are essentially using a stochastic mini-batch gradient.
\item 
Second, the size of the mini-batch is variable (depending
on the stopping criterion considered) and
it depends on some heuristics that estimates on the fly the ability
of the approximate gradient to retrieve the top entry of the true
gradient. 
\item Finally, one important component of our approach is
the importance sampling used in the stochastic mini-batch sampling.
This component theoretically helps in reducing the error of
the gradient estimation \cite{p14:_stoch_optim_impor_sampl}.
In the context of matrix multiplication approximation we used
for developing the randomized approach in Section \ref{sec:mvprod}, theoretical results of Drineas et al. \cite{drineas2006fast} have 
also shown there exists an importance sampling that minimizes
the expectation of the Frobenius norm of the matrix multiplication approximation.
Our experiments corroborate these results showing that,
compared to a uniform sampling, importance sampling clearly enhances the efficiency in retrieving the top entry of the true gradient. 
\end{itemize}
All these differences make our randomized algorithm not only clearly distinguishable
from  stochastic gradient approaches, but also harder to analyze. 
We thus defer for further work the theorical analyses of such
a stochastic adaptive-size mini-batch gradient coupled with an importance sampling approach. 
 
\subsection{Theoretical considerations}

Although a complete theoretical analysis of the algorithm is out
of the scope of this paper, an interesting property deserves to be
mentioned here. Note that, unlike stochastic gradient approaches,
our algorithm is built upon inexact gradients that hopefully
have the same minimum component as the true gradient. 
If this latter fact occurs along the iterations of the
FW or OMP algorithms, then all the properties (\emph{e.g} linear convergence,
 exact recovery property\ldots) of these
algorithms apply. Based on the probability of recovering
an exact minimum component of the gradient at each iteration,
we show below a bound on the probability of
our algorithm to recover at a given iteration
$K$ of OMP or FW, the same sequence of minimum components as
the one obtained with exact gradient. 

Suppose that at each iteration $t$ of OMP or FW, our algorithm for
estimating the minimum component correctly identifies
this component with a probability at least
$ 1 - P_t$ and there exists $\bar P$ so that 
$P_t> \bar P,\,\forall t \leq K$, then the probability
of identifying the correct sequence of minimum component
is at least
$
1- K \bar P.
$
We get this thanks to the following reasoning. 
Denote $B(t)=\{I^1= i_{e}^1,I^2= i_{e}^2,\ldots,I^t= i_{e}^t\}$
the event that our algorithm outputs the exact sequence of
minimum components up to iteration $t$, $I^t$ and $i_e^t$ being the
coordinates retrieved with the inexact and exact gradient. Similarly,
we note $A(t)=\{I^t= i_{e}^t\}$  the event  of retrieving, at iteration $t$, the correct component of the gradient. We assume that 
$\proba(B(0))=1$.  
\rev{
Formally, we are interested in lower-bounding the probability
of $B(K)$. By definition, we have
\begin{align*}\nonumber
\proba(B(K))&=\proba(A(K)|B(K-1)) \proba(B(K-1)) \\\nonumber
&=\proba(B(0)) \prod_{t=1}^K\proba(A(t)|B(t-1)). 
\end{align*}

Note that this equation captures all the time dependencies that occur during the FW or the OMP
algorithm. Since $\proba(A(t)|B(t-1)) \geq 1 - P_t$, we have
\begin{align*}\nonumber
\proba(B(K))&\geq  \prod_{t=1}^K(1-P_t)
\geq (1- \bar P)^K \geq (1- K \bar P)
\end{align*}
where in the last inequality, we used the fact
that $(1-u)^K \geq (1-Ku)$ for $0 \leq u \leq 1$. 
}

For instance,  in the \emph{successive halving} algorithm, we
have $P_t = 3 \log_2 d \cdot \exp\Big(- \frac{T}{8H_2(t)log_2 d}\Big)$,
where $H_2(t)$ is a iteration-dependent constant \cite{karnin2013almost}
and $T$ the number of pulls. Thus, if we 
define $\bar H$ so that $\forall t, \bar H \geq H_2(t)$, we
have $ \bar P = 3 \log_2 d \cdot \exp\Big(- \frac{T}{8 \bar H log_2 d}\Big)$ and
$$P(B(K)) \geq 1 -  3 K \log_2 d \cdot \exp\Big(- \frac{T}{8 \bar H log_2 d}\Big).
$$
We can see that the probability of our OMP or FW having the same behaviour as their exact counterpart decreases with the number $K$ of iterations
and the number $d$ of dimensions of the problems and increases with the number of pulls. By rephrasing this last equation, we also get the following
property. For $\delta \in [0,1]$, if the number $T$ of pulls is set so that at each iteration $t$,
$$
T \geq \left (\log \log_2 \frac{d}{\delta} + \log \frac{K}{\delta} + log \frac{3}{\delta}\right) 8 \bar H \log_2 d
$$
then, when using the \emph{successive halving} algorithm
for retrieving the extremum gradient component, an
inexact OMP or FW algorithm behaves  like the exact OMP or
FW with probability $1-\delta$.
This last property is another emphasis on the strength of 
our inexact gradient method compared to stochastic gradient descent approaches as it shows that with high probability, all the 
theoretical properties of OMP or FW (\emph{e.g} convergence, exact recovery of sparse signal) apply.

 \section{Numerical experiments}
\label{sec:expe}
In this section, we describe the experimental studies we
have carried out for illustrating the computational
benefits of using inexact gradient for sparsity-constraint
optimization problems.

\subsection{Experimental setting}

In order to illustrate the benefit of using inexact gradient
for sparse learning or sparse approximation, we have set up a simple sparse
approximation problem which focuses on the computational gain, and for
which a sparse signal has to be recovered by
the Frank-Wolfe, OMP or CoSaMP algorithm.

\revII{Note that sparse approximations are mostly used for
  approximation problems on overcomplete dictionary. This is the case
  in our experiments, where the dimension $d$ of the learning problem is
  in most cases larger than the number $n$ of samples.  We believe that
  if the signal or the image at hand to be approximated can be fairly
  approximated by representations for which fast transforms
are available, then it is better
  (and faster) to indeed used this representation and the fast
  transform. Sparse approximation problems as considered in
the sequel, mostly occur in overcomplete dictionary learning
problems. In such a situation, as the dictionary is data-driven, we believe that the approach we propose is
relevant.
 }

 The target
sparse signals are built as
follows. For a given value of the dictionary size
$d$ and a number $k$ of active elements in the dictionary, the true
coefficient vector $\w^\star$ is obtained as follows. The $k$ non-zero positions are chosen randomly, and their values
are drawn from a zero-mean unit variance Gaussian distribution,
to which we added $\pm 0.1$ according to the sign of the values.
The columns of the regression design matrix $\X \in \R^{n \times d}$ are drawn uniformly from the surface of a unit
hypersphere of dimension n. Finally, the target vector is obtained
as
$\y = \X \w^\star+ \e$,
where $\e$ is a random noise vector drawn from a Gaussian distribution with zero-mean and variance $\sigma_e^2$ determined from a given
signal-to-noise as
$
\sigma_e^2= \frac{1}{n}\|\X \w^\star\|^2 \cdot 10^{-SNR/10}.
$
Unless specified, the SNR ratio has been set to $3$.
For each setting, the results are averaged over 20 trials, and $\X$, $\w^\star$ and $\e$ are resampled at each trial.

The criteria used for evaluating and comparing the proposed approaches are the running time of the algorithms and
their ability to recover the true non-zero elements of $\w^\star$.
The latter is computed through the F-measure
between the support of the true coefficient vector $\w^\star$
and the estimated one $\hat \w$: $$
\text{F-meas}= 2 \frac{|\text{supp}_\gamma(\w^\star) \cup \text{supp}_\gamma(\hat \w)|}{
|\text{supp}_\gamma(\w^\star)|+|\text{supp}_\gamma(\hat \w)|}
$$
where, $\text{supp}_\gamma(\w) = \{j : |w_j| > \gamma\}$ is the support of vector $\w$ and $\gamma$ is a threshold used to neglect some
non-zero coefficients that may be obliterated by the noise.
In all our experiments, we have set $\gamma = 0.001$ which is small compared
to the minimal absolute value of a non-zero coefficient (0.1).

All the algorithms (Frank-Wolfe, OMP and CoSaMP) and exact and inexact gradient  codes have
been written in Matlab except for the \textit{successive reject} bandit  which as been written in C and transformed in a 
mex file. \rev{All computations have been run on each single core of an Intel Xeon E5-2630 processor clocked at 2.4 GHz in  a Linux machine with 144 Gb of memory.}

\subsection{Sparse learning using a Frank-Wolfe algorithm}

\begin{figure*}[t]
  \centering
 ~\hfill \includegraphics[width=7.3cm]{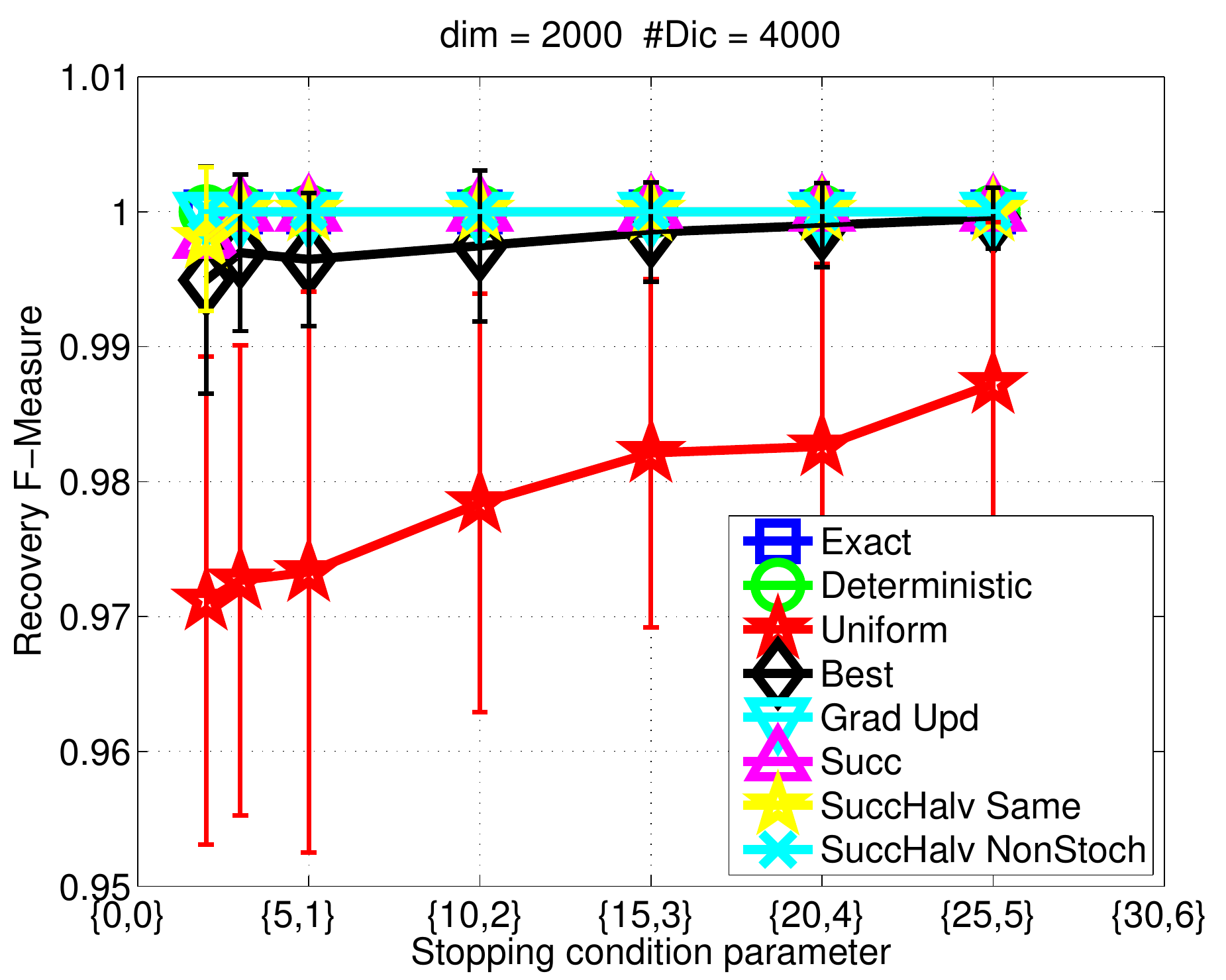}
\hfill
  \includegraphics[width=7.3cm]{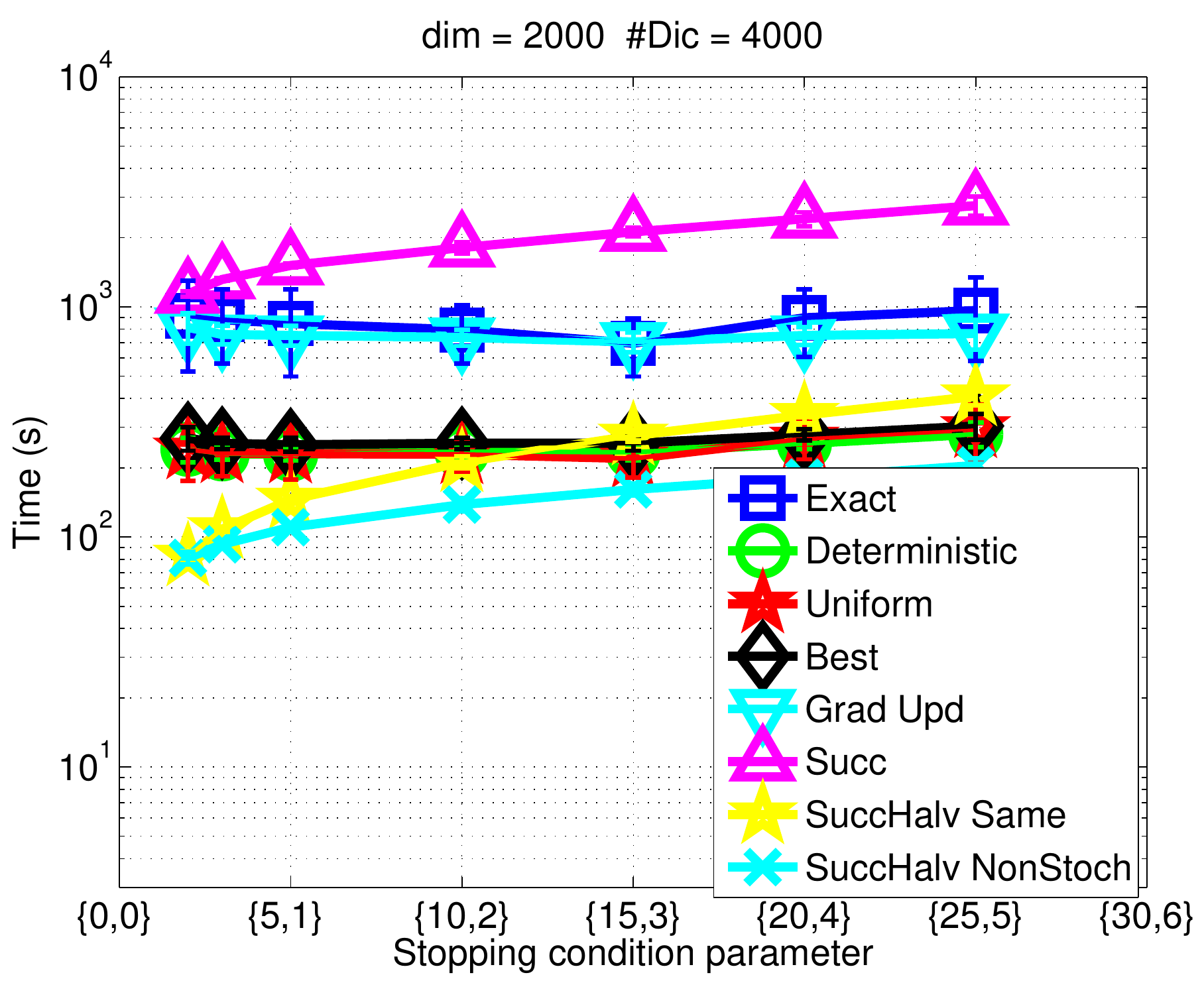}\hfill~
  \caption{Comparing vanilla Frank-Wolfe and inexact FW algorithms with different ways for computing the inexact gradient with $n=2000$, $d=4000$ and $k=50$. Performances are compared with increasing precision on the inexact computations. 
We report the exact recovery performance and the running time of algorithms.
\label{fig:fw}}
  \label{fig:fw2000}
\end{figure*}

\begin{figure*}[t]
  \centering
~\hfill  \includegraphics[width=6.5cm]{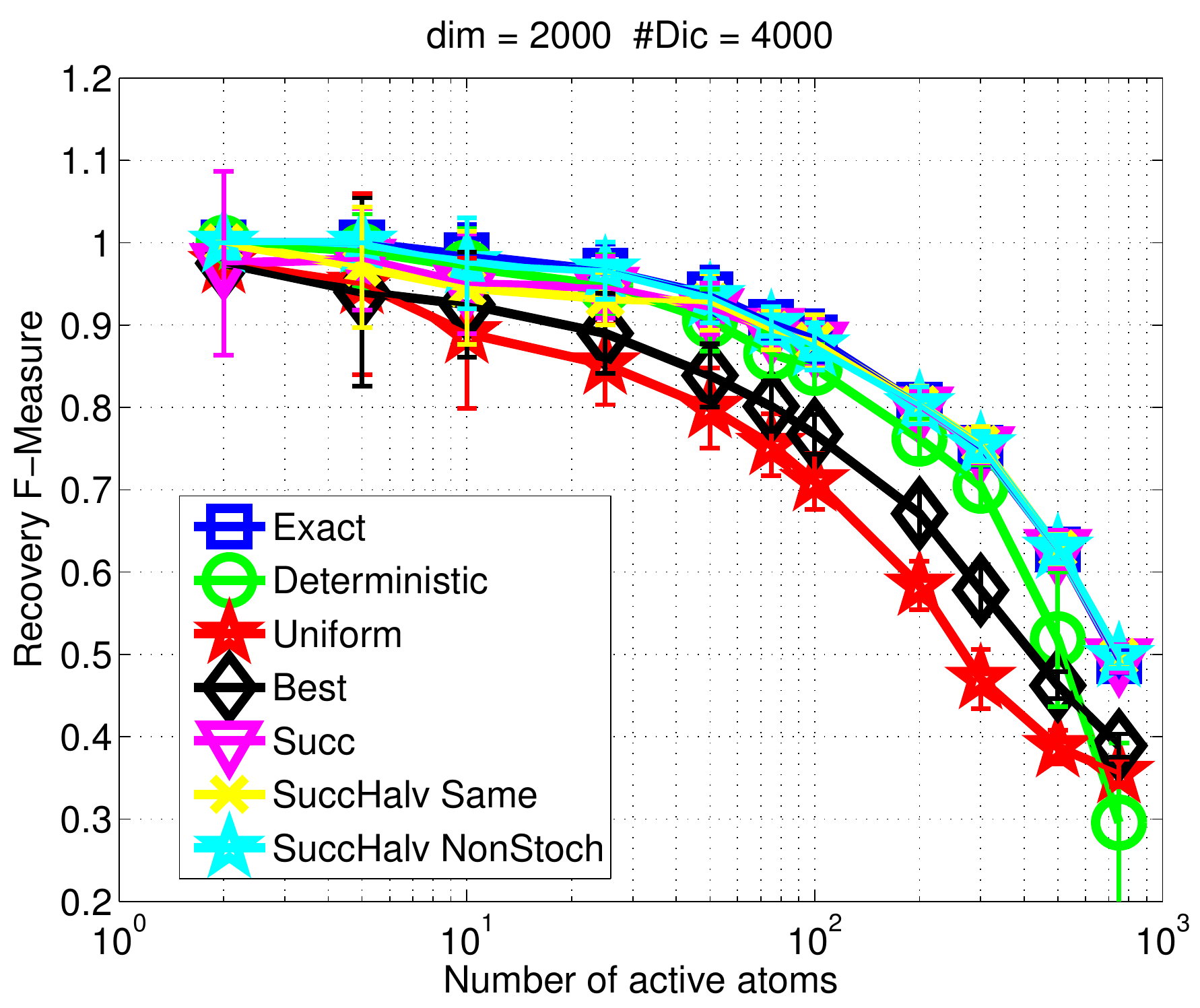}
\hfill
  \includegraphics[width=6.5cm]{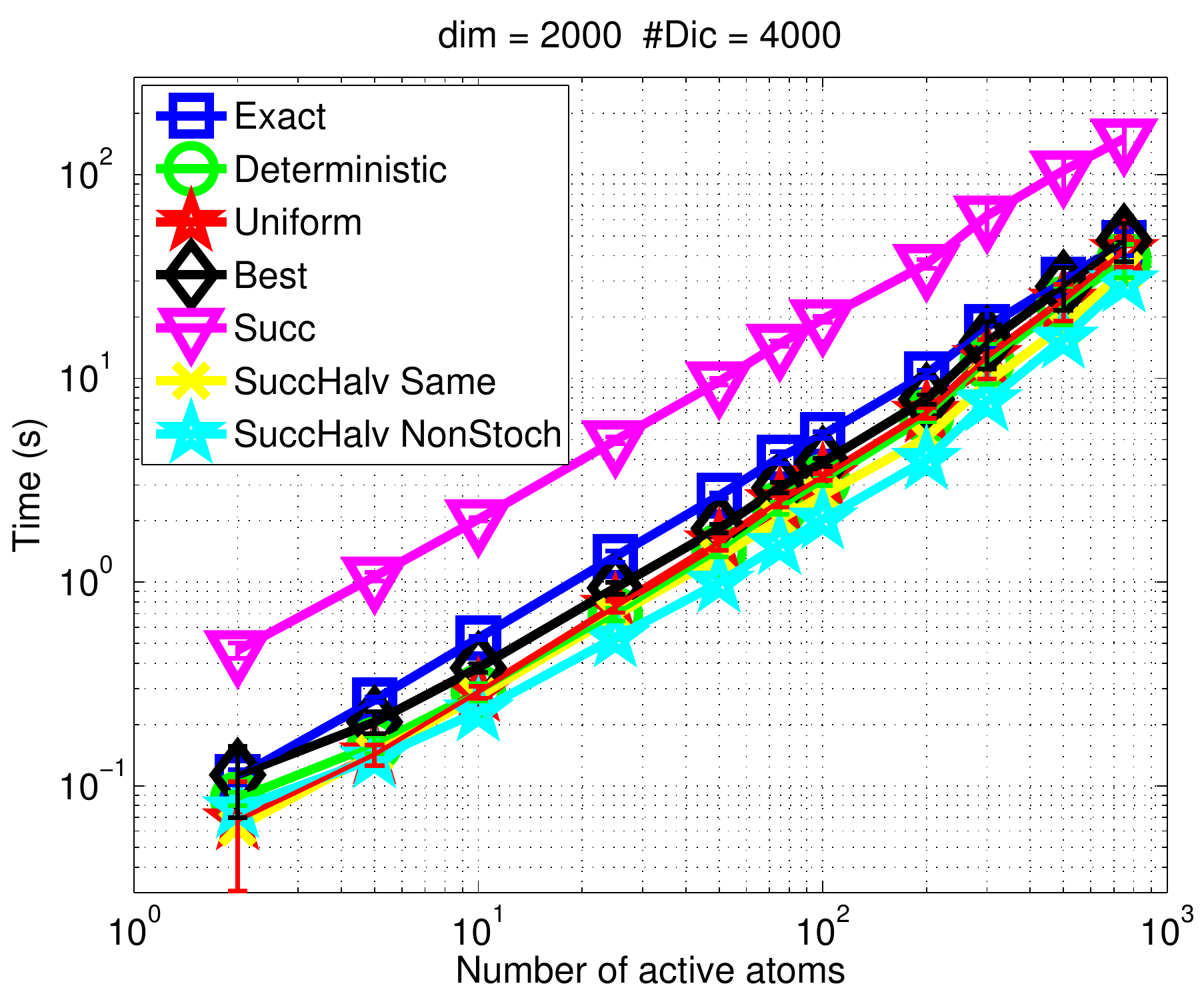} \hfill~\\
~\hfill  \includegraphics[width=6.5cm]{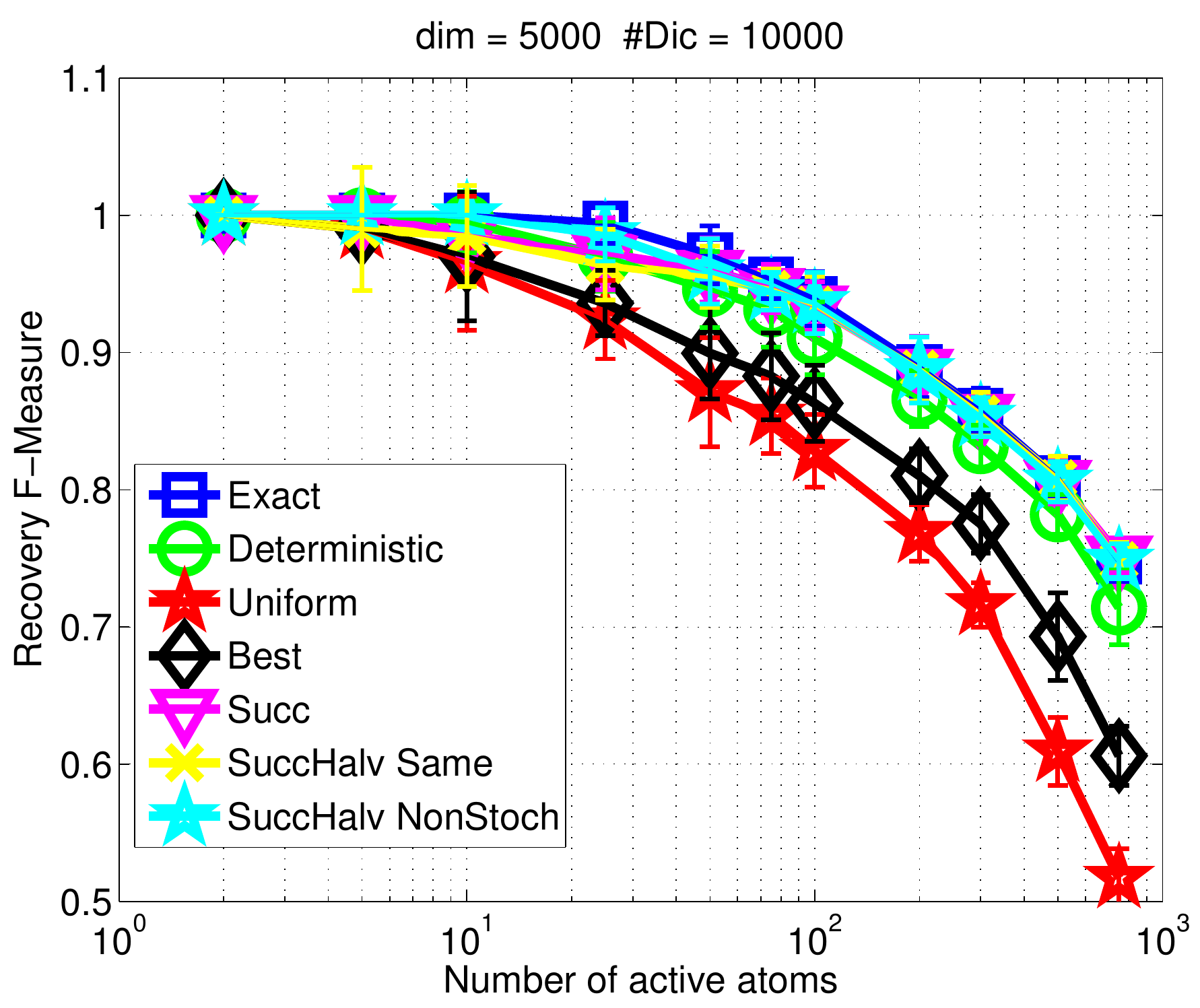}\hfill
  \includegraphics[width=6.5cm]{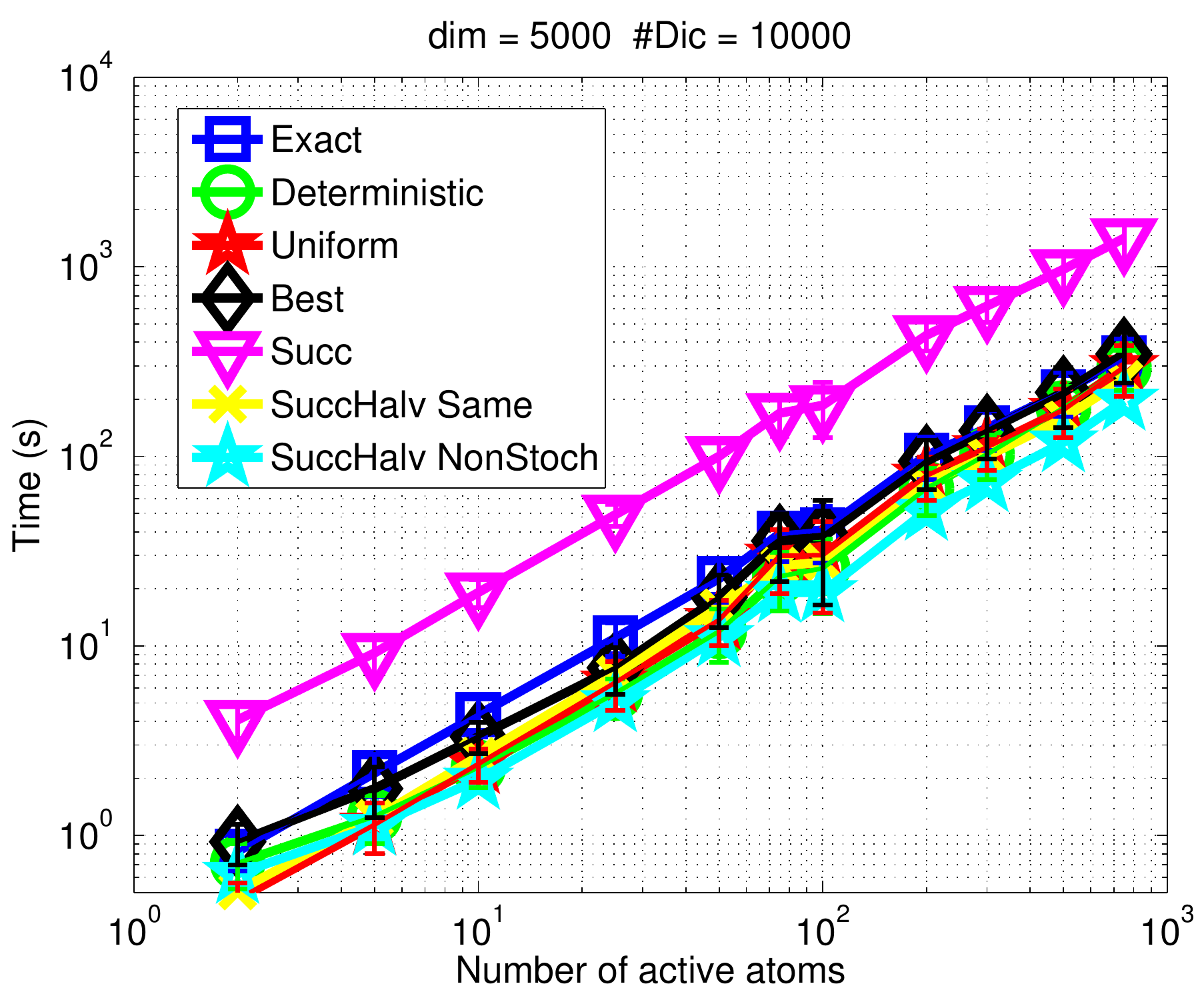} \hfill~\\
  \caption{Comparing OMP algorithm with different ways for computing the inexact gradient. The comparison holds for dictionary size with (top) $n=2000$, $d=4000$. (bottom) $n=5000$, $d=10000$ and for an exact recovery criterion (left) and
running time (right). }
  \label{fig:OMP}
\end{figure*}

For this experiment, the constraint set $C$ is the $\ell_1$ unit-ball and the loss function is $L(\w)=
\frac{1}{2}\|\y - \X\w\|_2^2$.
Our objectives are  two-folded:
\begin{enumerate}
\item analyze the capability of inexact gradient approaches to recover the true support and compare them to the FW algorithm,
\item compare the two stopping criteria for computing the inexact gradient : the \emph{stability condition} and the \emph{error bound condition}.
\end{enumerate}
In the latter, while this \emph{error bound condition} provides 
an adaptive condition for stopping --- recall that the parameter
$\epsilon$ in Equation (\ref{eq:errorcrit}) is determined automatically
through the data and the related curvature of the loss function ---, the
\emph{stability condition}  needs
a user-defined parameter $N_s$ for stopping the accumulation of the partial gradient.
In the same spirit, we use a fixed pre-defined  budget
of pulls  in the best-arm identification problem. This budget
is given as a ratio of $n\times d$. 
The exact gradient is computed using 
the accumulation strategy as given in Equation \eqref{eq:accumulation}
so as to make all running times comparable.
The maximum number of iterations for FW is set to $5000$.

Figure \ref{fig:fw} presents the results obtained for $n=2000$ samples, $d=4000$ dictionary elements and $k=50$ active
atoms. We depict the running time and recovery abilities
of the Frank-Wolfe algorithm with an exact gradient (\textbf{exact}), a greedy deterministic gradient sampling computation with a stability stopping criterion
(\textbf{deterministic}) and an error bound stopping criterion \textbf{(grad upb}), the randomized approach with an uniform sampling (\textbf{uniform}), and with a best probability sampling as
given in Equation \eqref{eq:best} (\textbf{best}), the successive reject bandit  (\textbf{succ}), the \emph{non-iid} successive halving approach 
with losses computed as given in Equation \eqref{eq:noniid} (\textbf{SuccHalvSame})
\rev{with a random uniform sampling} and the non-stochastic successive halving approach 
with losses computed as given in Equation \eqref{eq:banditnonstoch} (\textbf{SuccHalvNonStoch}) \rev{using a permutation $\tau$ that defines a decreasing ordering of the
$\|\x_i\| |r_i|$}. 

The figures depict the performances with respect
to  the stopping condition  parameter $N_s$ of the stability
criterion (the first value in the bracket) and the sampling budget of the bandit approach ($\frac{n\times d}{10} z$ where $z$ is the second value in the bracket). 
First, we can note that the deterministic approach used with any
stopping criterion and the non-stochastic successive halving approaches 
 are able to perfectly recover the exact support of the true vector $\w^\star$, regardless of the considered stopping criterion's value. 
Randomized approaches with uniform and best probability 
sampling nearly
achieve perfect recovery with an average F-measure of $0.975$
with the stability criterion $N_s$ equal to $5$ for the uniform
approach. When $N_s$ increases,
the performances of these two approaches also increase  but still fail to achieve perfect recovery.

From a running time point of view, the proposed approaches based on
greedy deterministic and randomized sampling strategies with stability
criterion and the successive halving strategies are faster than the
exact FW approach, the plain \emph{successive reject} method acting on
single entry of $\{x_{j,i} r_j\}$ and the deterministic method with
the \textit{error bound condition}. For instance, the greedy deterministic
approach
(green curve) achieves a gain
in running time of a factor $2$ with respect to the exact Frank-Wolfe
algorithm. Interestingly for the greedy deterministic approach and the
successive halving approaches, this gain is achieved without compromise on
the recovery performance.  For the randomized strategies,
increasing the stability parameter $N_s$ leads to a very slight
increase of running time, hence for these methods, a trade-off can
eventually be found.  When comparing bandit approaches, one can note
the substantial gain in performances that can be obtained by the
halving strategy, the \emph{non-iid} and \emph{non-stochastic}
strategies compared to \emph{successive reject}.  We conjecture that this higher computational running time
of the \emph{successive reject} algorithm is essentially due to
computational overhead needed for accessing each single matrix entry
$\X_{i,j}$ in memory while all other methods use slices of this matrix
(through the samples $\x_i$) and thus they can leverage on the chunk
of memory access. Best performances jointly  in recovery and
running times are achieved by the greedy deterministic and
the non-stochastic successive halving approaches.

When comparing the\textit{ stability }and the \textit{error bound}
stopping criteria, the latter one is rather inefficient. While
grounded on theoretical analysis, this bound is loose enough to be
non-informative.
\revII{Indeed, a careful inspection shows that 
the \textit{error bound} criterion accumulates about $5$ times more 
elements $r_i\x_i$ than the stability one before triggering. 
In addition, other computational overheads necessary for the bound estimation, make the approach just as efficient as  the exact Frank-Wolfe algorithm.
}

In summary, from this experiment, we can conclude that the 
non-stochastic bandit approach is the most efficient one. It
can achieve a gain in computation of about an order of magnitude
(the left most point in the Figure \ref{fig:fw2000}'s right panel) without compromising accuracy. 
The greedy deterministic approach with stability criterion performs
also very well but it is slightly less efficient. We can remark
that these two best methods both use the same strategy of gradient accumulation
based on decreasing ordering of $\|\x_i\|| r_i|$.

\subsection{Sparse Approximation with OMP}

Here, we evaluate the usefulness of using inexact gradient
in a greedy  framework like  OMP. 
The toy problem is similar to the one used above except
that we analyze the performance of the algorithm for
an increasing number $k$ of active atoms and two sizes of dictionary matrix $\X$ have been considered.
 
The same ways for computing the inexact gradient 
are evaluated and compared in terms of efficiency
and correctness to the true gradient in an OMP algorithm. 
For all sampling approaches, the stopping criterion for
gradient accumulation is based on the stability criterion with the
parameter $N_s$ adaptively set at $2\%$ of the number $n$
of samples.
For the successive reject bandit approach, the sampling budget has
been limited to $20\%$ of the number of entries (which
is $n\cdot d$ in the matrix $\X$.
In all cases, the stopping criterion for the OMP algorithm
is based on a fixed number of iterations and this
number is the desired sparsity $k$.

Results are reported in Figure \ref{fig:OMP}. They globally
follow the same trend as those obtained for the Frank-Wolfe algorithm.
First, note that in terms of support
recovery, when the number of active atoms is small,
the greedy deterministic approach  performs better than the
randomized sampling strategies. Bandit approaches perform similar to the greedy deterministic
method. As the number
of active atoms increases, the bandit approaches succeed better
in recovering the extreme component of the gradient while the deterministic
approach is slightly less accurate. Note that for any value of $k$, the randomized
strategies suffer more than the  other strategies for recovering 
the true vector $\w^\star$ support. From a running time point of view,
again, we note that the deterministic and \emph{non-iid} successive halving bandit approaches seem to be the most efficient methods. 
The gain in running time compared to the exact gradient OMP is slight
but significant while it is larger when 
comparing with the \textit{successive reject} algorithm.

\subsection{Sparse Approximation with CoSaMP}

\begin{figure*}[t]
  \centering
~\hfill\includegraphics[width=7.3cm]{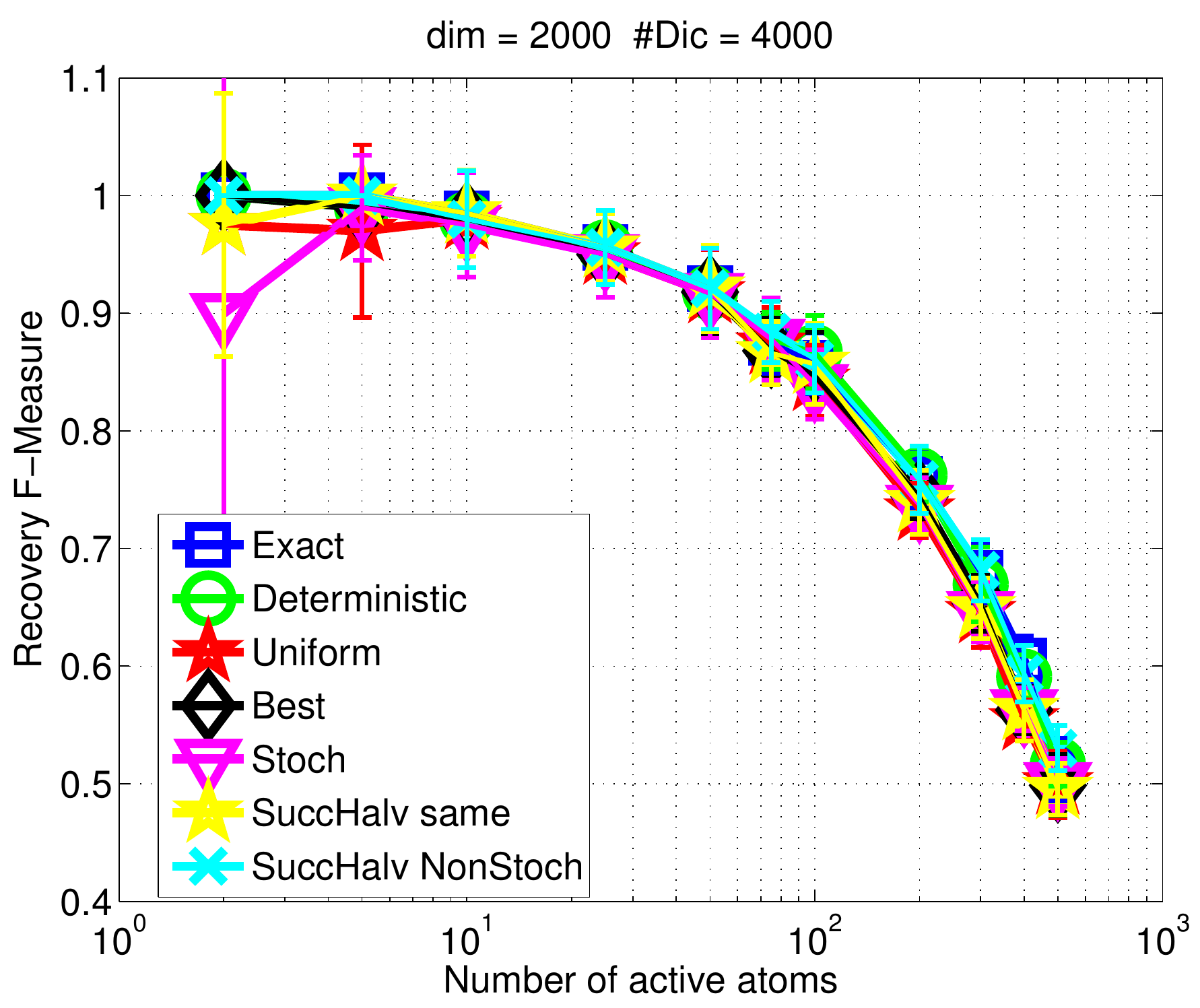}\hfill 
  \includegraphics[width=7.3cm]{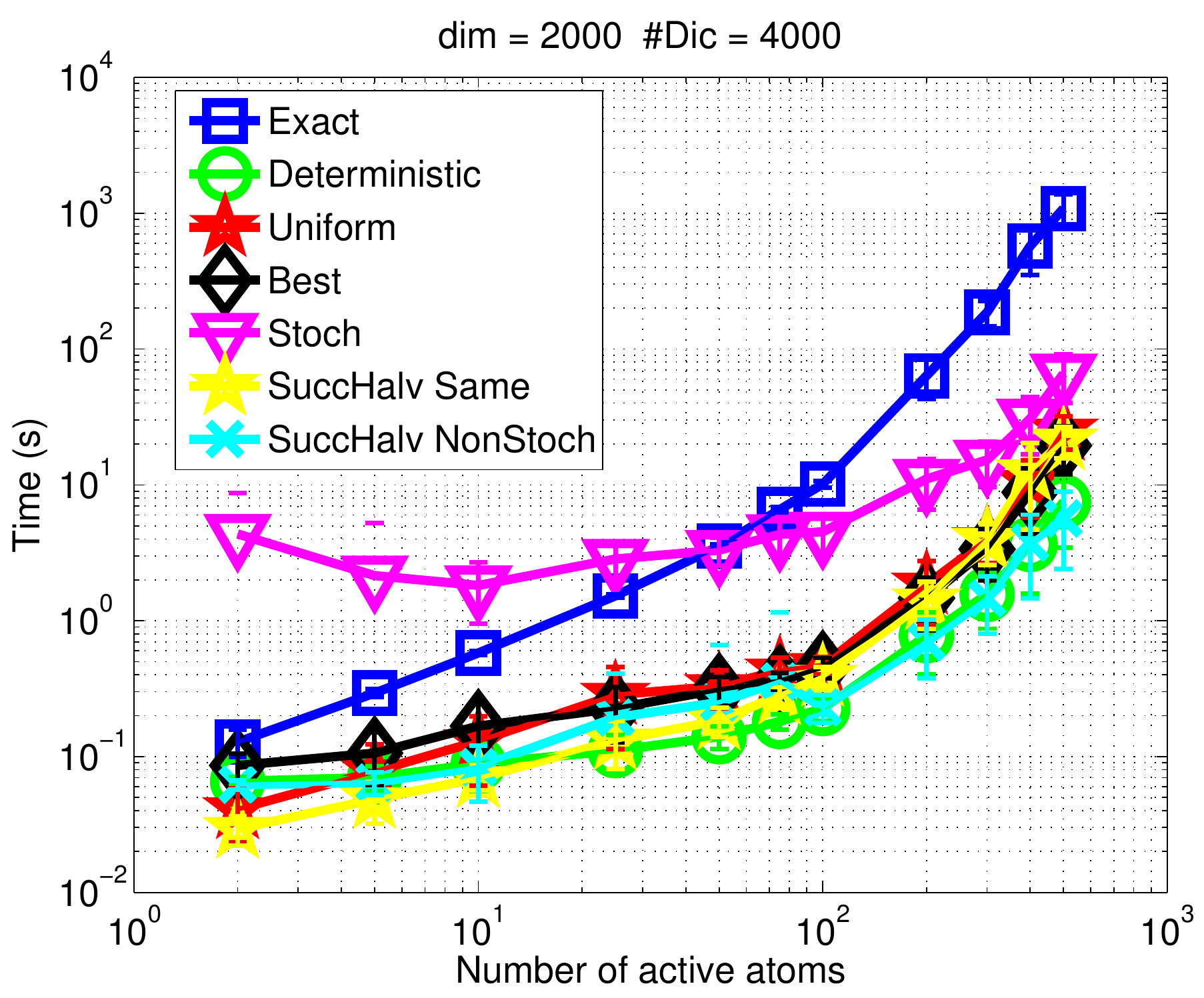}
\hfill~ \\
~\hfill  \includegraphics[width=7.3cm]{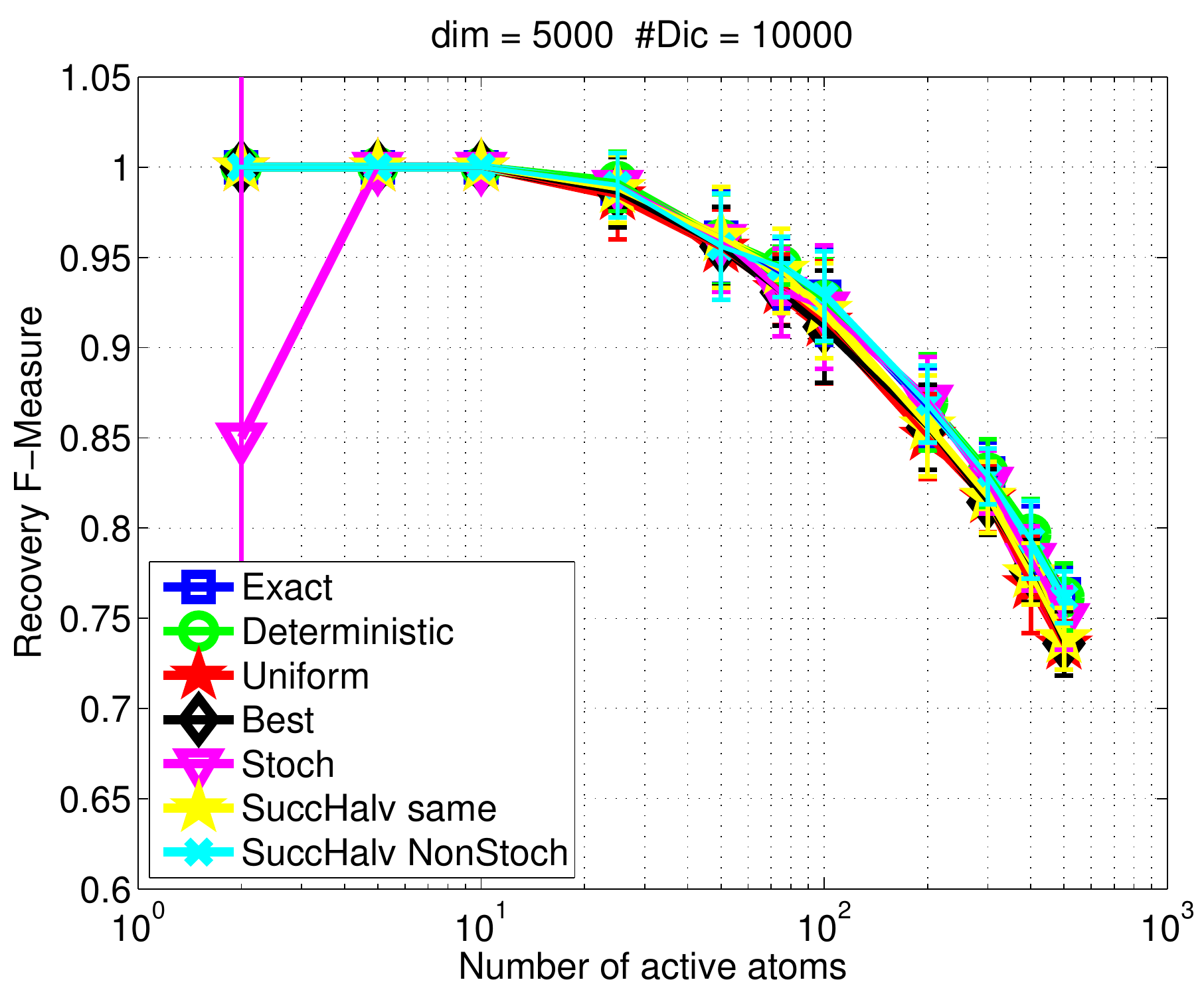}  \hfill
  \includegraphics[width=7.3cm]{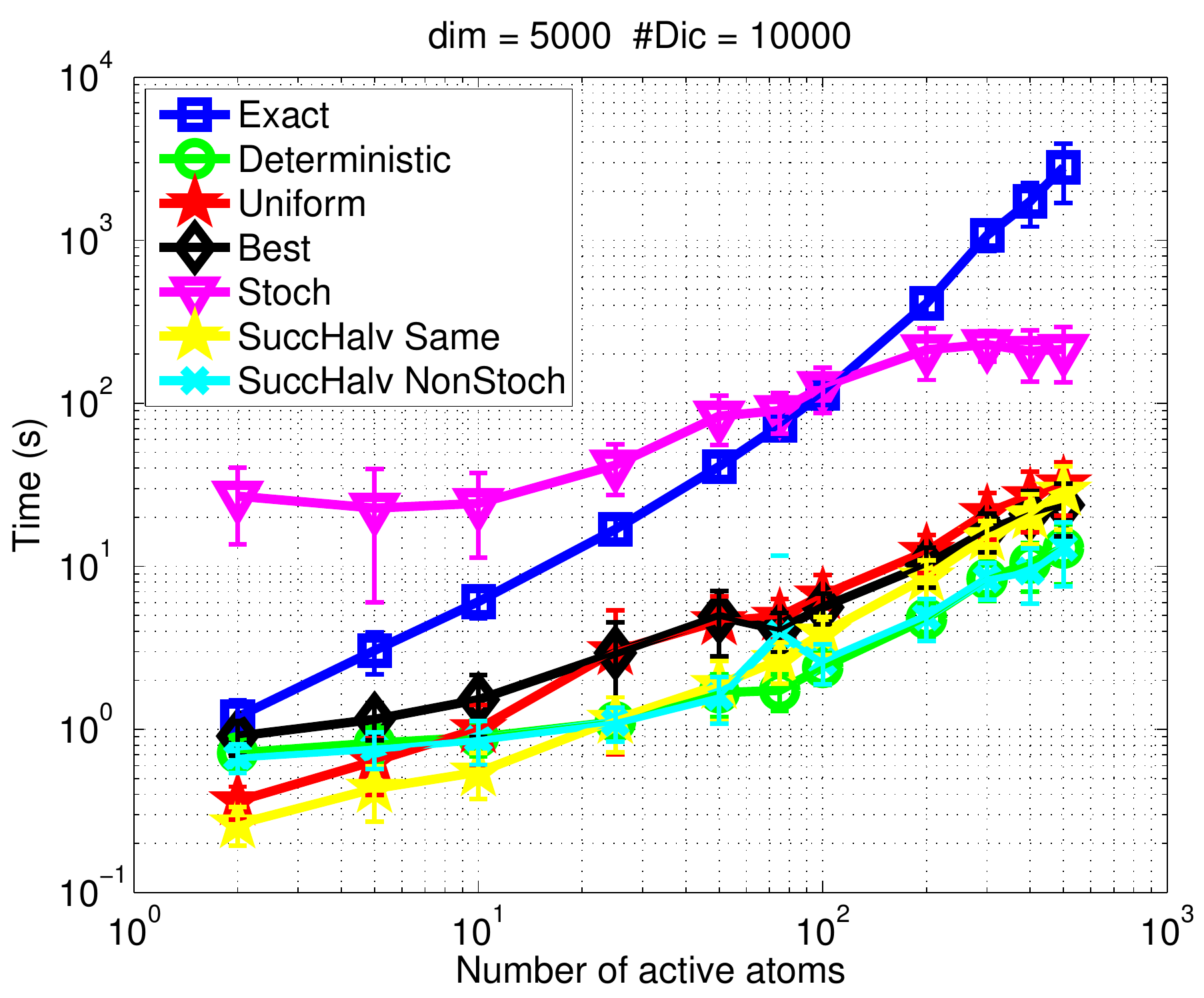} 
\hfill~
  \caption{Comparing CosAMP and  CosAMP algorithms with different ways for computing the inexact gradient.  The comparison holds for different dictionary size $n=2000$, $d=4000$ and $n=5000$, $d=10000$. We report  exact recovery performance and running time. }
  \label{fig:cosamp}
\end{figure*}

\begin{figure*}[t]
  \centering
 ~\hfill \includegraphics[width=7.3cm]{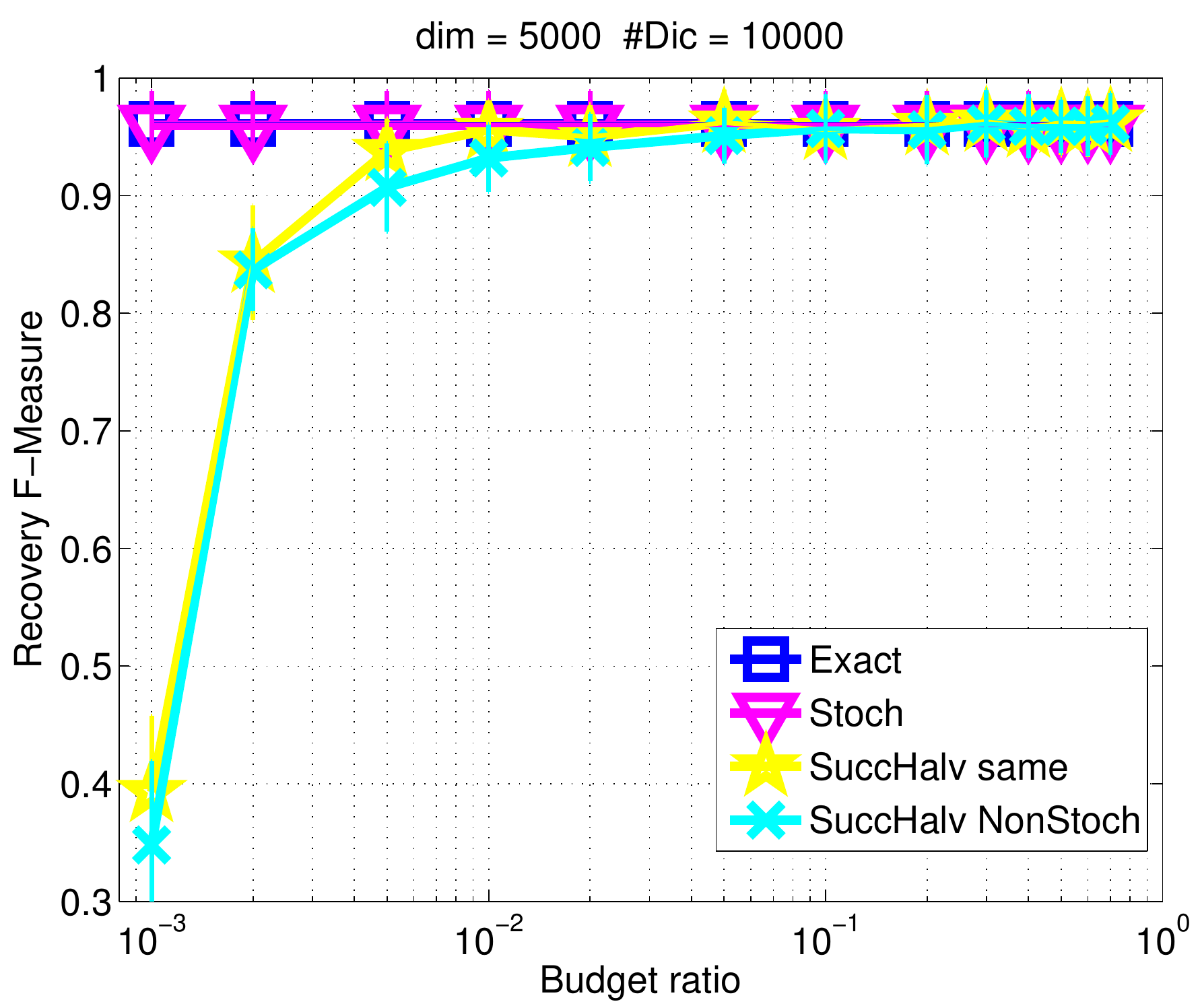} \hfill
  \includegraphics[width=7.3cm]{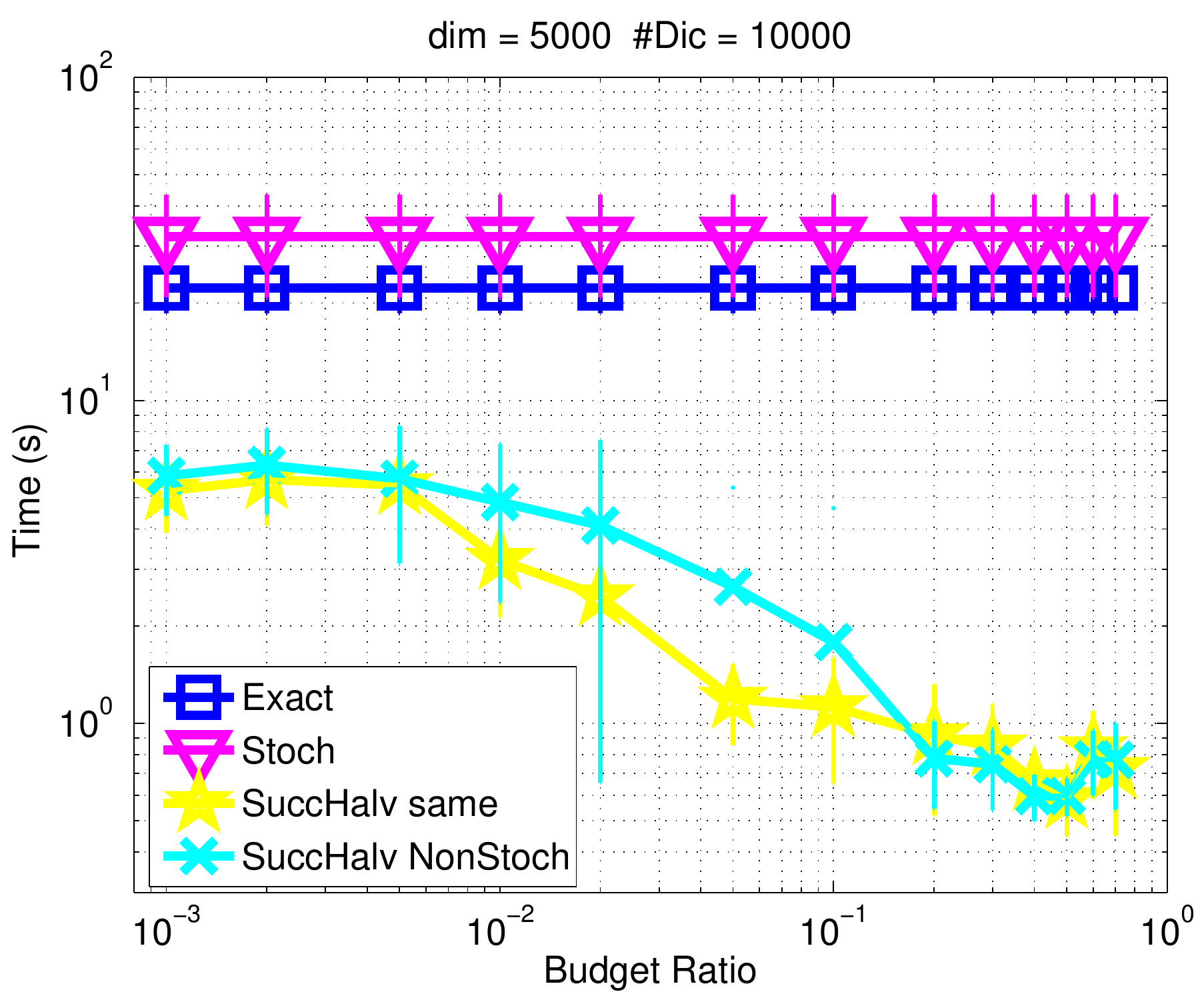}
\hfill~
  \caption{Analyzing the effect of the pull budget on the successive halving algorithm (left) recovery  F-measure and (right) computational running time. 
The pull budget is defined as the budget ratio times $n\cdot d$. Here
$n=5000$, $d=10000$ and $k=50$.}
  \label{fig:cosampbudget}
\end{figure*}

To the best of our knowledge, there are very few greedy algorithms
that are able to leverage from stochastic gradient. One
of these algorithms has been introduced in \cite{nguyen14:_linear}. 
In this experiment, we  want to evaluate the
efficiency gain achieved by our inexact gradient approach
compared to this stochastic greedy algorithm.
Our objective is to show that the approach we propose
is empirically significantly faster than a pure stochastic gradient approach.
For the different versions of the CoSaMP algorithm, we have
set the stopping criterion as follows. For the CoSaMP with
exact gradient approach, which serves only as a baseline
for computing the exact solution, the number of iteration is set to
the level of sparsity $k$ of the target signal.  A tolerance of the
residual norm is also used as a stopping criterion which should be
below $10^{-3}$.  Next, for the stochastic and the inexact gradient
CoSaMP versions, the algorithms were stopped when the norm of the
residual ($\y - \X\w$) became smaller than $1.001$ times the one
obtained by the exact CoSaMP or when a maximal number of
iterations. Regarding gradient accumulation, the stopping criterion we
choose is based on the stability condition with the parameter $N_S$
set dynamically at $2\%$ of the number of samples.  For the bandit
approaches, we have fixed the budget of pulls at $0.2 n \times d$.

Note that for the CoSaMP algorithm, we do not look for the top entry
of the gradient vector but for the top $2k$ entries as such, we have thus
straightforwardly adapted the \textit{successive halving} algorithm to
handle such a situation.

Figure \ref{fig:cosamp} presents the observed results. Regarding
support recovery, we remark that all approaches achieve performances
similar to the exact CoSaMP.  When
few active atoms are in play, we can note that sometimes, the
stochastic approach of \cite{nguyen14:_linear} fails to
recover the support of $\w^\star$.  This occurs seldom but it
happens regardless of the dictionary size we have experimented
with.

From a running time perspective, the results show that the proposed
approaches are highly more efficient than the exact gradient approach
and interestingly, they are faster than a pure gradient stochastic
approach. One or two orders of magnitude can be gained depending on
the level of sparsity of the signal to be recovered.  This observation
clearly depicts the trade-off that occurs in sparsity-constrained
optimization problems in which the gradient computation and an
approximation problem on a limited number of atoms are the major
computational burdens (Lines 3 and 5 of Algo~\ref{algo:gp}). Indeed in
a stochastic gradient approach, inexact gradient computations are very
cheap but more approximation problems to be solved may be needed for
achieving a desired accuracy. In the approaches we propose, the
inexact gradient computation is slightly more expensive but we somehow
``ensure'' that it provides the correct information needed by the
CoSaMP algorithm. Hence, our approaches need less approximation
problems to be solved, making them more efficient than a stochastic
gradient approach.
 
When comparing the efficiency of the proposed algorithms, 
the approach based on non-stochastic successive halving
and the greedy deterministic approach are the most efficient
especially as the number of active atoms grows. \\

In the second experiment, for the \emph{successive halving} algorithms, we analyze the effect of the pull budget  on the running time and on the recovery
performance. We consider the following setting
of the problem: $n=5000$, $d=10000$ and $k=50$. 
Results are given in Figure \ref{fig:cosampbudget}. 
We note that a budget ratio varying from $0.05$ and $0.7$ 
allows a good compromise between ability
of recovering the true vector and gain in computational time,
particularly for the non-stochastic successive halving method.
As the budget of pulls decreases, both algorithms fail more
frequently in recovery and in addition, the computational gain
substantially reduces. This experiment suggests that one should not
be too greedy and should allow sufficient amount of pulls.
A budget ratio of $0.2$ or $0.3$  for the bandit algorithms
seems to be a good rule of thumb according to our experience.

\begin{figure*}[t]
  \centering
~\hfill\includegraphics[width=7.4cm]{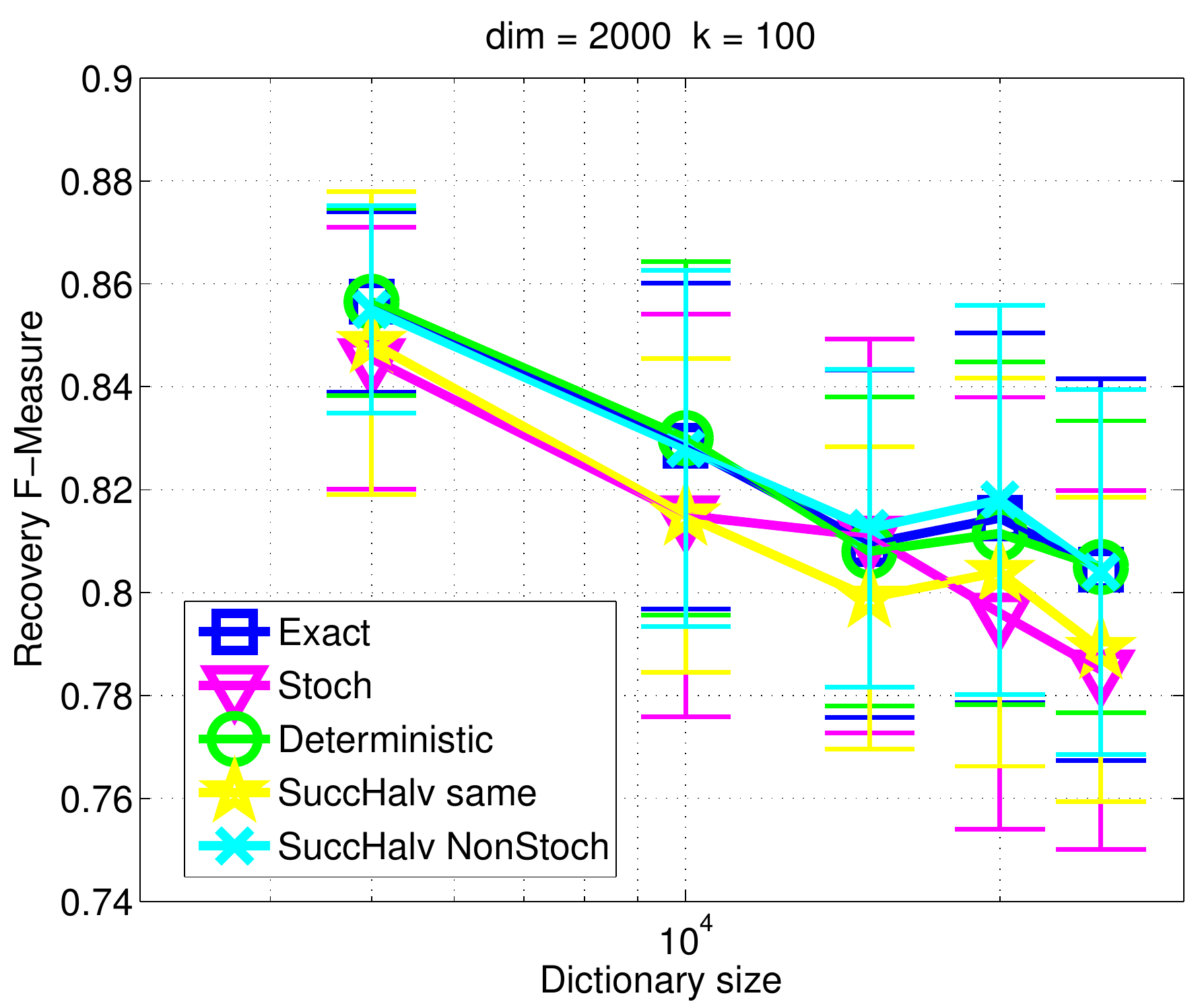}\hfill 
 \includegraphics[width=7.4cm]{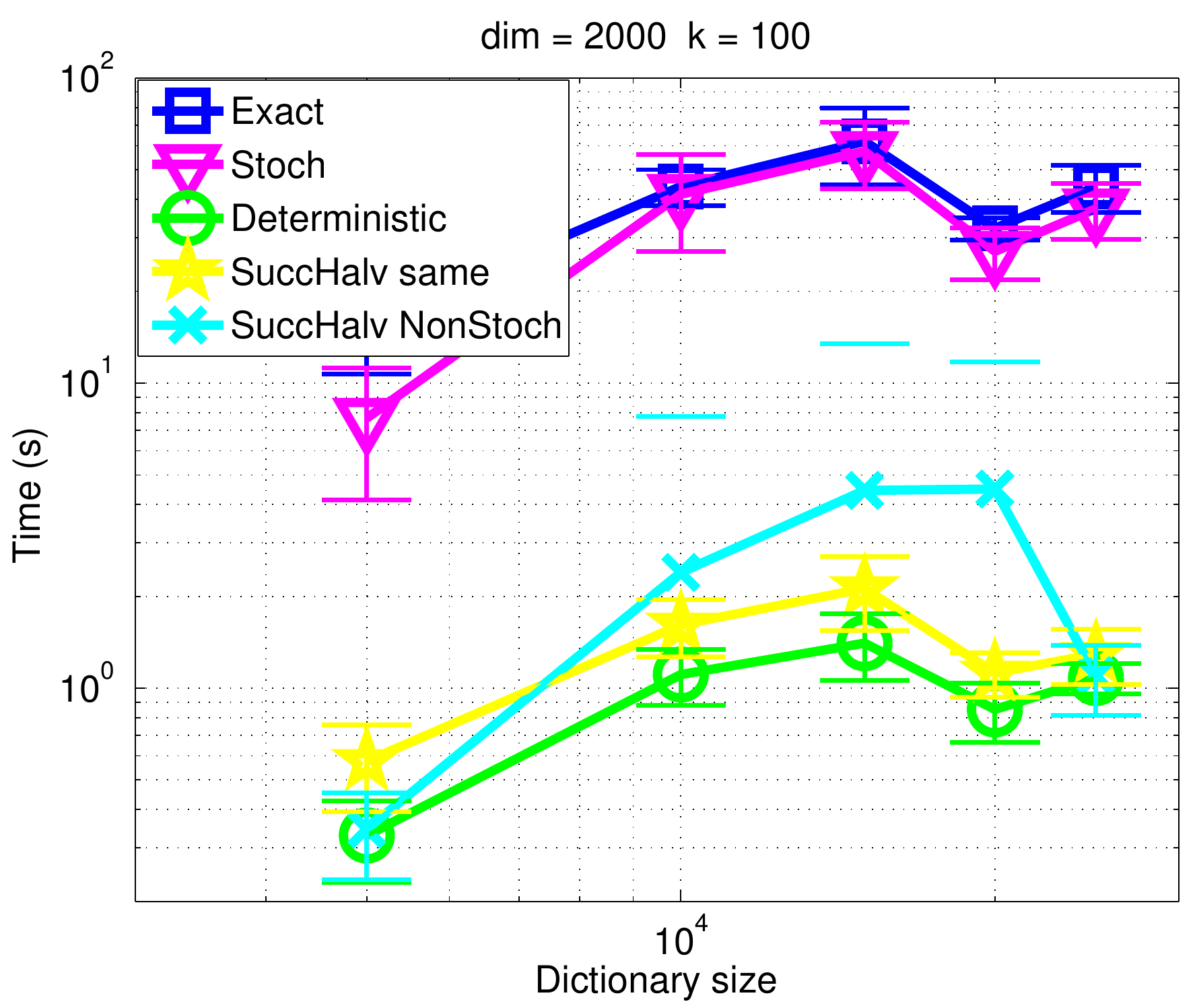}
\hfill~ \\
~\hfill\includegraphics[width=7.4cm]{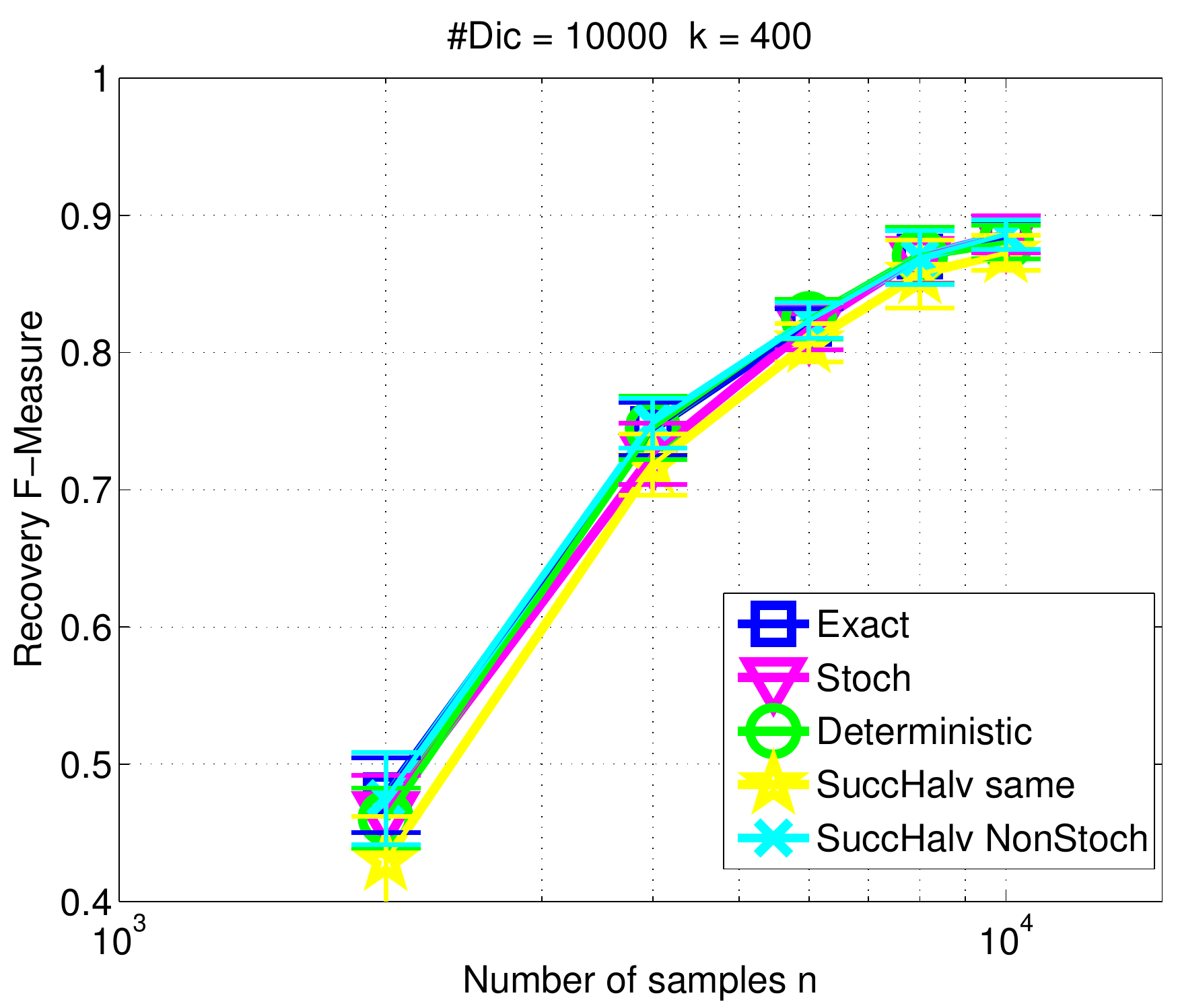}\hfill 
  \includegraphics[width=7.4cm]{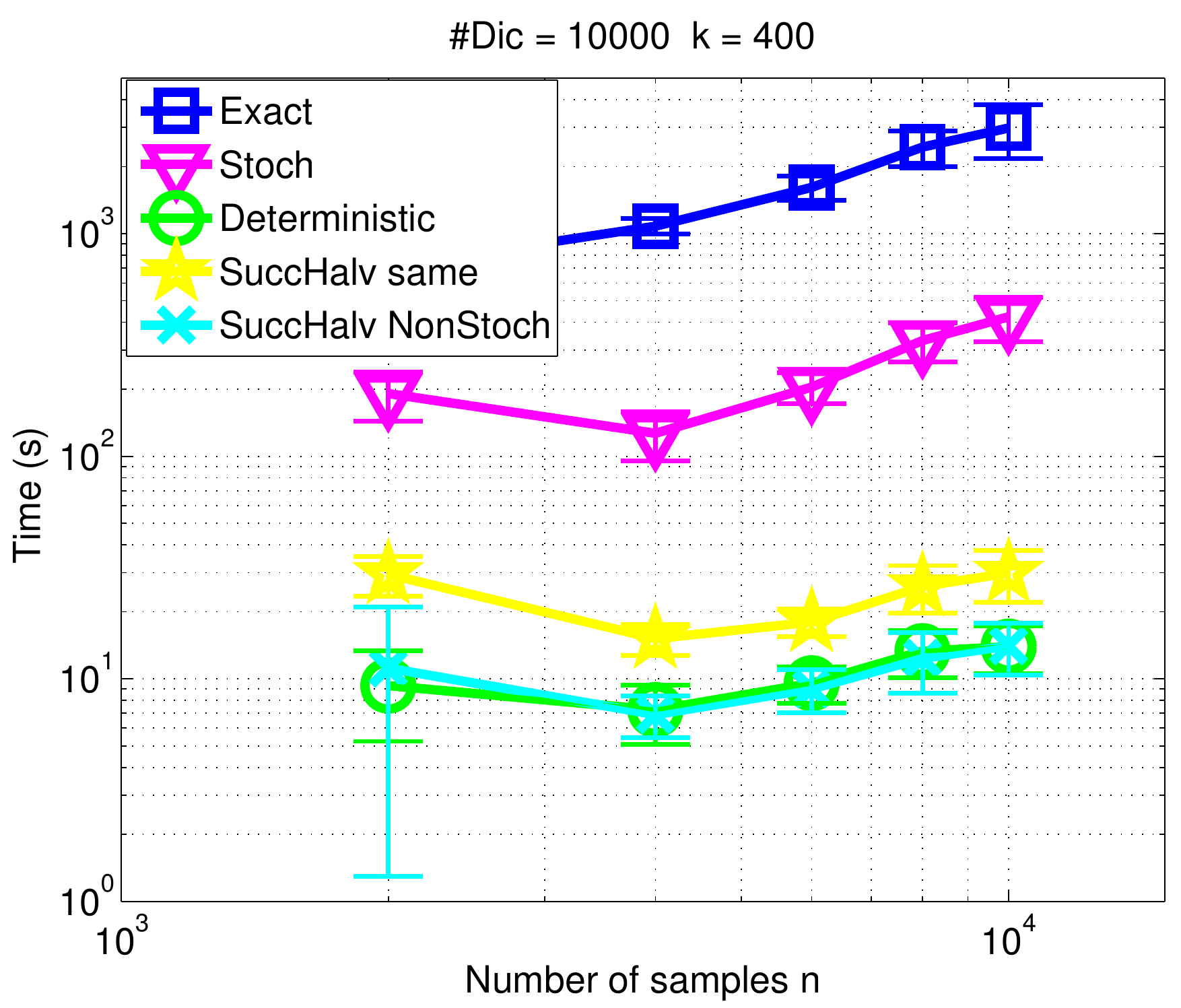}
\hfill~\\
  \caption{(top left and top right) Evaluating how  the recovery capability behaves and how  the computation time scales with the number
of dictionary elements (with $n=2000$ and $k=100$).
 (bottom left and bottom right) Evaluation of the
same criteria with respect to the number of samples (with $d=10000$ and $k=400$). }
  \label{fig:cosampcomplexity}
\end{figure*}

\rev{Our last experiment with CoSaMP demonstrates how the
running time and the support recovery performance behave
with an increasing number $n$ of samples and afterwards with an increasing 
number of dictionary elements $d$. We have restricted
our comparison to the exact and stochastic CoSaMP, 
and the CoSaMP variants based on the successive
halving bandit algorithms and the greedy deterministic one (which
are the most efficient among those we propose).
The experimental setup, the stopping criterion for the CoSaMP algorithm as well as the stopping criterion
for the gradient accumulation and pull budget are the same as above.
Results are depicted in Figure \ref{fig:cosampcomplexity}. As a
sanity check, we note that recovery performances are almost
similar for all algorithms with slightly worse performances
for the stochastic CoSaMP and the \emph{non-iid} bandit algorithm
based CoSaMP.

The computational time results show that all algorithms globally
follow the same trend as the number of dictionary atoms
or the number of samples increase. 
Recall that the computational complexity for the gradient computation
is $O(nd)$. For the bandit approaches, we use a fixed
budget of pulls dependent on $nd$ to compute the
inexact gradient. Similarly, for the greedy deterministic approach,
the number of accumulation (and the stability criterion) is proportional to the number $n$ of samples and thus the gradient computation
is a constant factor of $nd$. Hence, our findings, illustrated on 
Figure  \ref{fig:cosampcomplexity}, are somewhat natural
since the  main differences of
running time essentially come from  a constant factor.
This factor is highly dependent on the problem but according
to our numerical experiments, a ten-fold factor computational gain can be 
expected in many cases.
}
\subsection{Application to audio data}

We have compared the efficiency of the  approaches we propose on a real signal processing application.
The  audio dataset we use is the one considered by Yaghoobi et al. \cite{yaghoobi09:_diction_learn_for_spars_approx}. This dataset is composed of an audio sample recorded from a BBC radio session which plays classical music. From that audio sample, $8192$ pieces of signal have
been extracted, each being composed of $1024$ time samples. Details about
the dataset can be found in \cite{yaghoobi09:_diction_learn_for_spars_approx}. 
From this dataset, we have learned  $2048$ dictionary atoms 
using the approach described in \cite{rakotomamonjy12:_direc}.
\rev{Our objective is to perform sparse approximation of each of the $8192$ audio pieces  over the $2048$
dictionary atoms using CoSaMP and we want to evaluate the running time and the approximation quality of a 
CoSaMP algorithm using an exact gradient computation (\textbf{Exact}), a stochastic
gradient CoSaMP algorithm (\textbf{Stoch $k$}) and the CoSaMP variants with
inexact gradient computations as we propose. The approximation error
is measured as $\frac{\|\y - \hat \y\|_2}{\|\y\|_2}$ where $\y$ and
$\hat \y$ are respectively the true audio piece and its CoSaMP-based approximation. 
We thus want to validate that our approaches achieve similar approximation
performance than CoSaMP while being faster. For all
algorithms, the number of CoSaMP iterations is fixed
to the sparsity pattern, here fixed to $k=10$. Note that
for the stochastic gradient approach, we have also
considered a version with more iterations (\textbf{Stoch $3k$}) . 
}
Results are gathered in Table \ref{tab:error} and they are
obtained as the averaged performance when approximating
 all the $8192$ pieces of audio signal in the dataset.
We can see that the inexact approaches we introduce lead
to the best compromise between approximation error and running
time. \rev{For instance, our \emph{successive halving} algorithms 
achieve similar approximation errors than the exact CoSaMP but they are $3$ times
faster.} At the contrary, the stochastic gradient CoSaMP approaches are efficient
but lack in properly approximating target audio pieces.

\begin{table}[t]
  \caption{Approximation performance results and running time for CoSaMP and
variants. $\y$ and $\hat \y$ respectively depicts the signal and
its resulting approximation. Results are averaged over the approximation of $4500$ signals. 
\emph{Stoch $3k$} denotes the stochastic gradient algorithm that
used $3k$ iterations. }
 \label{tab:error}
  \centering
  \begin{tabular}[h]{l|cc}
   \hline 
Approaches & $\frac{\|\y - \hat \y\|}{\|\y\|}$ & Time (s) \\\hline\hline
 exact  & 0.376 $\pm$ 0.22 & 0.164 $\pm$ 0.01 \\ 
  stoch $k$  & 0.670 $\pm$ 0.13 & 0.026 $\pm$ 0.00 \\ 
  stoch $3k$  & 0.570 $\pm$ 0.17 & 0.076 $\pm$ 0.01 \\ 
  uniform  & 0.351 $\pm$ 0.21 & 0.133 $\pm$ 0.02 \\ 
  deterministic   & 0.361 $\pm$ 0.22 & 0.187 $\pm$ 0.02 \\  
  SuccHalvSame  & 0.371 $\pm$ 0.22 & 0.059 $\pm$ 0.01 \\ 
  SuccHalvNonStoch  & 0.374 $\pm$ 0.22 & 0.064 $\pm$ 0.01 \\\hline
  \end{tabular}
\end{table}

\subsection{Benchmark classification problems}

\begin{table*}[t]
\begin{center}
\caption{Comparing performances of CoSaMP and its variants with approximate
gradients on real-world high-dimensional classification problems. 
(top) statistic summary of datasets. (middle) Accuracy of the decision function. (bottom) Running time (in seconds)
of the learning algorithms.
 \label{tab:real}}
\begin{tabular}{rccccc}
&\multicolumn{5}{c}{Datasets}  \\\hline
\multicolumn{1}{c}{Information}& \multicolumn{1}{c}{ohscal} &\multicolumn{1}{c}{classic} & \multicolumn{1}{c}{la2} & \multicolumn{1}{c}{hitech} & \multicolumn{1}{c}{sports} \\\hline
\multicolumn{1}{c}{n} & 11162 & 7094 &3075 & 2301 & 8580 \\
\multicolumn{1}{c}{d} & 11465  &41681 & 31472 & 10080 & 14866\\
\hline
\vspace{0.1cm}
\\
&\multicolumn{5}{c}{Datasets}  \\\hline
\multicolumn{1}{c}{Algorithms}& \multicolumn{1}{c}{ohscal} &\multicolumn{1}{c}{classic} & \multicolumn{1}{c}{la2} & \multicolumn{1}{c}{hitech} & \multicolumn{1}{c}{sports} \\\hline\hline
CoSAMP & 82.93 $\pm$ 1.9& 83.46 $\pm$ 1.8& 81.20 $\pm$ 2.5& 76.29 $\pm$ 3.7& 93.07 $\pm$ 1.3\\\hline
Stoch & 56.16 $\pm$ 19.3& 63.90 $\pm$ 22.6& 70.24 $\pm$ 5.1& 11.63 $\pm$ 11.1& 40.74 $\pm$ 22.8\\\hline
Stoch $3k$ & 36.92 $\pm$ 26.2& 52.58 $\pm$ 26.8& 69.76 $\pm$ 5.5& 8.42 $\pm$ 3.1& 32.54 $\pm$ 20.4\\\hline
Determ. & 83.20 $\pm$ 1.6& 82.35 $\pm$ 1.3& 77.29 $\pm$ 2.3& 75.98 $\pm$ 2.9& 92.30 $\pm$ 1.4\\\hline
Uniform & 77.50 $\pm$ 1.7& 77.17 $\pm$ 1.2& 78.98 $\pm$ 2.9& 68.88 $\pm$ 3.6& 92.33 $\pm$ 1.4\\\hline
HalvingSame & 81.39 $\pm$ 1.5& 82.57 $\pm$ 1.5& 81.54 $\pm$ 2.6& 75.16 $\pm$ 2.1& 93.23 $\pm$ 1.2\\\hline
HalvingNonStoch & 81.90 $\pm$ 1.5& 82.97 $\pm$ 1.7& 79.91 $\pm$ 2.6& 76.49 $\pm$ 3.8& 93.21 $\pm$ 1.0\\\hline
\vspace{0.1cm}
\\
&\multicolumn{5}{c}{Datasets}  \\\hline
\multicolumn{1}{c}{Algorithms}& \multicolumn{1}{c}{ohscal} &\multicolumn{1}{c}{classic} & \multicolumn{1}{c}{la2} & \multicolumn{1}{c}{hitech} & \multicolumn{1}{c}{sports} \\\hline\hline
CoSAMP & 1257.89 $\pm$ 306.3& 563.51 $\pm$ 208.5& 401.36 $\pm$ 222.1& 56.42 $\pm$ 19.4& 998.63 $\pm$ 385.3\\\hline
Stoch & 3.12 $\pm$ 0.8& 1.05 $\pm$ 0.7& 2.98 $\pm$ 2.0& 0.69 $\pm$ 0.3& 4.13 $\pm$ 1.8\\\hline
Stoch $3k$ & 7.93 $\pm$ 1.8& 2.73 $\pm$ 1.6& 8.50 $\pm$ 5.4& 1.90 $\pm$ 0.8& 12.08 $\pm$ 4.8\\\hline
Determ. & 323.65 $\pm$ 76.6& 142.06 $\pm$ 62.3& 209.70 $\pm$ 130.5& 41.24 $\pm$ 16.3& 353.04 $\pm$ 129.9\\\hline
Uniform & 416.74 $\pm$ 102.2& 189.14 $\pm$ 74.4& 150.28 $\pm$ 89.2& 21.00 $\pm$ 7.7& 348.30 $\pm$ 117.0\\\hline
HalvingSame & 17.50 $\pm$ 5.2& 13.20 $\pm$ 12.5& 14.68 $\pm$ 15.2& 1.48 $\pm$ 0.6& 11.94 $\pm$ 6.3\\\hline
HalvingNonStoch & 17.61 $\pm$ 5.4& 11.74 $\pm$ 11.9& 13.54 $\pm$ 15.9& 1.37 $\pm$ 0.5& 10.57 $\pm$ 6.5\\\hline
\end{tabular}
\end{center}
\end{table*}
\rev{
We have also benchmarked our algorithms on real-world high-dimensional learning
classification problems. These datasets are frequently used  for
evaluating sparse learning problems \cite{gong2013jieping,rakotomamonjy2015dc}
and more details about them can be found in 
these papers.  
Here, we considered CoSaMP as a learning algorithm and our objective is to validate the fact that the approaches we propose for computing approximate gradient are able to speed up computation time while achieving the same level of accuracy as the exact gradient.
For the approximate gradient computations, we have considered the stability
criterion with $N_S=\frac{n_{train}}{20}$  ($n_{train}$ being the number of training examples) for the deterministic and randomized approaches, and we have set the budget as $0.2 n_{train} \times d $ for the bandit approaches.

The protocol we have set up is the following. Training and test sets are obtained by
randomly splitting the dataset in a $80\%-20\%$ fold. For model selection,
the training set is further split in two sets of equal size. The parameter we have 
cross-validated is the number of non-zero elements $k$ in $\w$. It has been
selected among $10,50 ,100, 250$ so as to maximize the accuracy on the
validation set. This value of $k$ has also been used as the maximal number
of iterations for all the algorithms except for the stochastic ones. 
For these, we have reported accuracies and running times for
a number of maximal iterations of $k$ and $3k$. 

Results averaged over $20$ replicas of the training
and test sets are reported in Table \ref{tab:real}. Stochastic
approaches fail in learning a relevant decision function.
We can note that our deterministic and randomized 
approaches are more efficient than the exact CoSaMP but are less  accurate. On the other hand, our bandit approaches achieve
nearly similar accuracy to CoSaMP while being at least $30$ times 
faster.
}

\section{Conclusions}
\label{sec:conclusion}
The methodologies proposed in this paper aim at 
accelerating sparsity-constrained optimization algorithms.
This is made possible thanks to the key observation that, at each
iteration, only the component of the gradient
with smallest or largest entry is needed, instead of the full gradient.
By exploiting this insight, we proposed greedy algorithms,
randomized approaches and bandit-based best arm identification methods
for estimating efficiently this top entry. 
Our experimental results show that the bandit and the greedy approaches 
seem to be the most efficient methods for this estimation. 
\rev{Interestingly, the bandit approaches come with  guarantees
that, given a sufficient number of draws, this top entry
can be retrieved with high-probability. 

Future works will be geared towards gaining further theoretical
understandings on the good behaviour of the greedy approach,
linking the number of iterations needed for the Frank-Wolfe algorithm
to converge, with the quality of the gradient approximation
in the greedy and randomized approaches, 
 analyzing 
the role of the importance sampling in the randomized methods. 
In addition, we plan to explore how this work can be extended
to an online and/or distributed computation setting.
}

\bibliographystyle{IEEEbib}

\end{document}